\documentclass{article}

\PassOptionsToPackage{numbers, compress}{natbib}

\usepackage[preprint]{neurips_2026}
\usepackage{enumitem}
\usepackage{wrapfig}
\usepackage[htt]{hyphenat}
\usepackage{multirow}
\usepackage{booktabs}
\usepackage{siunitx}
\usepackage{float} % 导言区
\usepackage{xcolor}
\usepackage{listings}
\usepackage{tabularx}
\usepackage{hyperref} % only needed if not already included

\usepackage[table]{xcolor} % for \rowcolor and \columncolor
\usepackage{array}
\usepackage{pgfplots}
\usepackage{booktabs}
\usepackage{caption}
\usepackage{subcaption}
\usepackage{siunitx} % 对齐小数点可选

\usepackage{booktabs}     % \toprule, \midrule, \bottomrule
\usepackage{makecell}     % 多行表头 \makecell
\usepackage{adjustbox}    % \begin{adjustbox}{max width=\textwidth}
\usepackage{graphicx}     % 如果要旋转文字可以用 \rotatebox（可选）

\usepackage{booktabs}
\usepackage{subcaption}

\usepackage[utf8]{inputenc} % allow utf-8 input
\usepackage[T1]{fontenc}    % use 8-bit T1 fonts
\usepackage{hyperref}       % hyperlinks
\usepackage{url}            % simple URL typesetting
\usepackage{booktabs}       % professional-quality tables
\usepackage{amsfonts}       % blackboard math symbols
\usepackage{nicefrac}       % compact symbols for 1/2, etc.
\usepackage{microtype}      % microtypography
\usepackage{xcolor}         % colors

\definecolor{reconblue}{RGB}{0, 173, 216}

\colorlet{headerblue}{blue!10}

\usepackage[most]{tcolorbox}
\usepackage{xcolor}

\definecolor{myBG}{RGB}{245,245,245}

\tcbset{
  promptbox/.style={
    colback=myBG,
    colframe=black,
    arc=2mm,
    boxrule=0.5pt,
    left=2mm,
    right=2mm,
    top=1mm,
    bottom=1mm,
    breakable
  }
}

\newcommand{\dashdash}{\texttt{-\kern0pt-}}

\title{KaliBench: A Fine-Grained Benchmark for Cybersecurity Tool Use on Kali Linux with Runtime-Free Verifiable Rewards}

\usepackage{risys-title}

\RisysTitle{KaliBench: A Fine-Grained Benchmark for
Cybersecurity Tool Use on Kali Linux with
Runtime-Free Verifiable Rewards}
  \RisysAuthor[1,*]{Pengfei Li}
  \RisysAuthor[1,*]{Naufal Suryanto}
  \RisysAuthor[1]{Sicheng Zhang}
  \RisysAuthor[1,2]{Muzammal Naseer}

  \RisysAffil[1]{Khalifa University}
  \RisysAffil[2]{University of Western Australia}
  \RisysAffil[*]{Equal contribution}
\RisysProjectpage{https://risys-lab.github.io/KaliBench/}
\RisysGithub{https://github.com/RISys-Lab/KaliBench}
\RisysAbstract{
LLMs are increasingly applied to cybersecurity workflows, where they are expected to translate analysts' intent into tool invocations. However, existing evaluations focus on knowledge-based assessments or end-to-end agentic tasks, and do not directly measure LLMs’ ability to generate executable commands for real-world cybersecurity tools. This gap is critical because cybersecurity operations rely on strict command-line interfaces (CLIs), where minor syntax errors, incorrect flag--value bindings, or argument misordering can invalidate execution. We introduce KaliBench, a fine-grained benchmark and dataset for natural-language--to--CLI translation on Kali Linux, comprising 8,504 query--command pairs spanning 1,642 tools across 23 capability dimensions and 5 security phases. KaliBench is constructed via a manuscript-grounded pipeline with deterministic canonicalization and alias-aware evaluation, enabling precise and reproducible assessment of tool selection and argument construction. To ensure both semantic correctness and practical executability, we develop a multi-stage verification pipeline that combines LLM-based validation, sandboxed terminal execution, and human-in-the-loop refinement. Building on these fine-grained, deterministic signals, KaliBench further enables runtime-free verifiable rewards for training. Across three evaluation modes and 24 configurations of general-purpose and security-focused open-weight models, no open-weight model exceeds 42\% exact-command accuracy in the unrestricted setting, highlighting the difficulty of accurate CLI-based cybersecurity tool use without explicit tool hints. We further show that supervised fine-tuning and reinforcement learning with verifiable rewards derived from KaliBench significantly improve an 8B model and achieve performance comparable to a 685B MoE model.}

\begin{document}

% \maketitle

\RisysMakeTitle

\begin{figure*}[!b]
\centering
% \vspace{-15pt}
% \fbox{\rule{0pt}{4.5cm}\rule{0.95\textwidth}{0pt}}
\includegraphics[width=\linewidth]{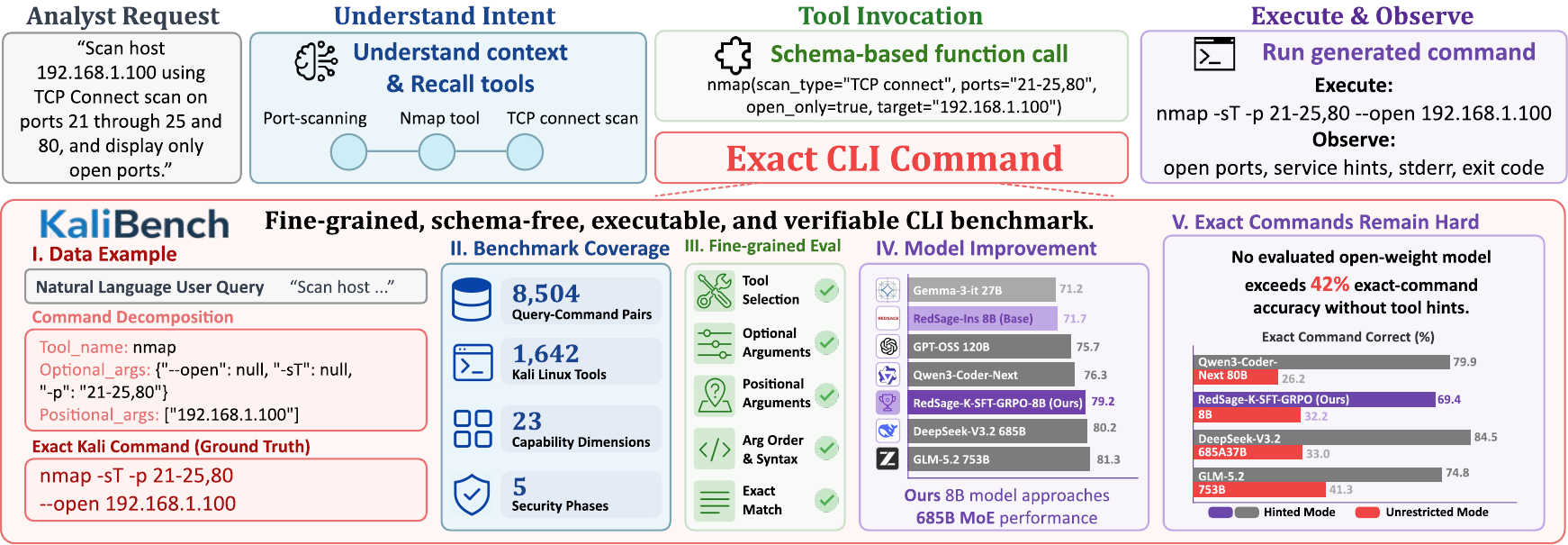}
\caption{\textbf{Why KaliBench Matters?} KaliBench enables fine-grained, verifiable evaluation and runtime-free reward training for translating analyst intent into executable Kali commands.
}
\label{fig:teaser}
\end{figure*}

% \end{abstract}

\section{Introduction}

% \NS{TODO: Update Intro to update datasets statistics, include training pipeline, and new results.}

% \begin{figure}[h]
%     \centering
%     \includegraphics[width=0.4\columnwidth]{image/windmill_11790.pdf}
%     \caption{\textbf{KaliBench} organizes 23 Kali Linux capability dimensions across the full cyber chain: from reconnaissance to defensive forensics. The inner arcs visualize tool counts for each security phase, and the outer arc visualizes a more detailed tool count for the 23 security dimensions. Each counted tool is associated with multiple query-command pairs.
%     }
%     \label{fig:kalibenchDim}
%     % \vspace{-3.75mm}
% \end{figure}

% LLMs in cybersecurity and gaps in existing evaluation.
Large language models (LLMs) are increasingly adopted in cybersecurity workflows, ranging from interactive assistants to autonomous agents \cite{llm4cybersecurity_survey}. Early benchmarks in this domain primarily focus on \emph{knowledge-based} assessment, testing models’ understanding of cybersecurity concepts via multiple-choice \cite{cybermetric, liu2024cyberbench} or structured-answer questions \cite{secbench, bhusal2024secure}. 
More recent benchmarks move toward \emph{end-to-end agentic} evaluation, measuring model performance in complex environments such as autonomously solving CTF-style challenges \cite{shao2024nyu-ctf, zhang2024cybench}. While these benchmarks provide valuable insight into what LLMs know and how they behave in integrated workflows, they do not directly evaluate whether models can generate correct invocations of concrete cybersecurity tools in isolation.
% , tool invocation is not evaluated in isolation. %

% \begin{figure*}[!h]
% \centering
% \vspace{-15pt}
% % \fbox{\rule{0pt}{4.5cm}\rule{0.95\textwidth}{0pt}}
% \includegraphics[width=\linewidth]{image/teaser-v2.pdf}
% \caption{\textbf{Why KaliBench Matters?} KaliBench enables fine-grained, verifiable evaluation and runtime-free reward training for translating analyst intent into executable Kali commands.
% % \SC{The old ver. looks like a AI-gen figure, the new version remove the execution workflow and highlight our direct NL-to-CLI task (left), and thus give more place to the experiments part.}
% }
% \label{fig:teaser}
% \end{figure*}

% CLI-centric nature of cybersecurity tooling.
This limitation is consequential because modern security operations rely heavily on large, heterogeneous command-line toolchains. Kali Linux, a widely used platform among cybersecurity practitioners, provides access to over 2,000 pre-installed security tool packages \cite{kali_tools}. In practice, analysts must translate high-level goals (e.g., ``scan this subnet for SMB misconfigurations'') into precise command-line invocations of tools such as \textit{Nmap} for scanning, \textit{Metasploit} for exploitation, and \textit{Volatility} for memory forensics. Although LLMs with function-calling capabilities aim to automate this process, CLIs are unforgiving: minor errors in argument order, flag spelling, alias misuse, or tool misinterpretation can invalidate execution. This motivates a central question: \emph{Can current LLMs reliably translate natural-language security requests into exact, executable CLI commands?}

% % Function-calling benchmarks and schema-free gap.
% While function-calling benchmarks for LLMs have matured rapidly, evaluating cybersecurity tool invocation poses additional challenges. Benchmarks such as API-Bank \cite{li2023api-bank}, StableToolBench \cite{guo2024stabletoolbench}, ToolSandbox \cite{lu2025toolsandbox}, APIGen \cite{liu2024apigen}, and BFCL \cite{patil2025bfcl} formalize tools using JSON schemas or HTTP endpoints that explicitly define supported functionality and parameters. In contrast, cybersecurity tooling is dominated by CLIs, such as the Kali Linux tools, where enumerating hundreds of tool schemas in a model prompt is impractical for most LLMs under context-length and efficiency constraints. As a result, NL-to-CLI evaluation is effectively \emph{schema-free}: models must rely on internal knowledge or compact task-specific hints, and must generate syntactically correct, executable commands rather than populate structured fields.

% Function-calling benchmarks and schema-free gap.
While function-calling benchmarks for LLMs have matured rapidly \cite{li2023api-bank, guo2024stabletoolbench, lu2025toolsandbox, liu2024apigen, patil2025bfcl}, evaluating cybersecurity tool invocation poses distinct challenges. Existing benchmarks primarily assume schema-defined tools, in which functionality and parameters are explicitly specified via JSON schemas or APIs. In contrast, cybersecurity tooling is dominated by CLIs, such as the Kali Linux toolset, where enumerating hundreds of tool schemas in a model prompt is impractical in typical deployment settings due to context length limitations. As a result, Natural-Language-to-CLI (NL--to--CLI) evaluation is effectively \emph{schema-free}: models must infer appropriate tools and arguments and generate syntactically correct, executable commands rather than populate structured fields.

% While function-calling benchmarks for LLMs have matured rapidly \cite{li2023api-bank, guo2024stabletoolbench, lu2025toolsandbox, liu2024apigen, patil2025bfcl}, evaluating cybersecurity tool invocation poses distinct challenges. Existing benchmarks primarily assume schema-defined tools, in which functionality and parameters are explicitly specified via JSON schemas or APIs. In contrast, cybersecurity tooling is dominated by CLIs, such as the Kali Linux toolset, where enumerating hundreds of tool schemas in a model prompt is impractical in typical deployment settings. As a result, NL-to-CLI evaluation is effectively \emph{schema-free}: models must infer appropriate tools and arguments and generate syntactically correct, executable commands rather than populate structured fields.

% Our proposal
To address this gap, we develop \textbf{KaliBench}, a fine-grained benchmark for NL--to--CLI 
cybersecurity tool use on Kali Linux. As shown in Fig.~\ref{fig:teaser}, KaliBench maps analyst intent to exact CLI commands, decomposing tool selection and argument construction for fine-grained, verifiable evaluation. It organizes tools across capability dimensions and security phases and is constructed via a manuscript-grounded pipeline. To ensure data quality, we employ a multi-stage verification process combining LLM-based validation, sandboxed execution, and human-in-the-loop refinement, producing semantically correct and executable commands.

% To address this gap, we develop \textbf{KaliBench}, a fine-grained benchmark for NL--to--CLI 
% % \SC{First time use this abbr., we need mention the full name, also the usage of - and -- is inconsistent in this paper} \NS{Will revise the intro after finishing rewriting the experiment, discussion, and conclusion}. 
% cybersecurity tool use on Kali Linux. As shown in Fig.~\ref{fig:teaser}, KaliBench organizes tools across capability dimensions and security phases and is constructed via a manuscript-grounded pipeline. To ensure data quality, we employ a multi-stage verification process combining LLM-based validation, sandboxed execution, and human-in-the-loop refinement, producing commands that are both semantically correct and 
% executable. 
% \NS{Finalize after Fig.1 completed.}

KaliBench provides fine-grained supervision over tool selection and argument construction. We evaluate models under three settings, \textit{unrestricted}, \textit{restricted}, and \textit{hinted}, which vary the degree of tool information provided. This structured formulation further enables \emph{runtime-free verifiable rewards}, allowing training without command execution. Using this framework, we evaluate 24 configurations of general-purpose and
security-focused open-weight models and find that performance remains limited, with argument construction as the primary bottleneck. We further show that Supervised Fine-Tuning (SFT) and Reinforcement Learning with Verifiable Rewards (RLVR) improve an 8B model and achieve performance competitive with substantially larger models.
% Contributions
Our main contributions are summarized as follows:
\begin{itemize}[leftmargin=*]
    \item \textbf{A fine-grained benchmark for schema-free NL-to-CLI evaluation.} KaliBench is, to the best of our knowledge, the first benchmark for evaluating LLMs on cybersecurity tool invocation in schema-free CLI settings, comprising 8,504 query--command pairs over 23 capability dimensions and five security phases, split into 5,000 evaluation and 3,504 training samples after deduplication.
    
    \item \textbf{A verified and executable dataset.} We construct a high-quality dataset through a multi-stage verification pipeline that ensures both semantic correctness and practical executability.
    
    \item \textbf{Runtime-free verifiable rewards.} We show that deterministic CLI structure enables verifiable reward signals without runtime execution, supporting scalable reinforcement learning.
    
    \item \textbf{Fine-grained evaluation and training gains.} We evaluate over 20 LLMs under three settings and identify argument construction as the primary bottleneck. Training with supervised fine-tuning and reinforcement learning using KaliBench rewards improves an 8B model's mean Total Score across the three evaluation modes by 7.5 percentage points.
\end{itemize}

\section{Related Work}
% =========== Revised Naufal ============
\paragraph{LLMs and Cybersecurity Benchmarks.}
% Early LLM evaluation focuses on general tasks (MMLU, HELM)
Early evaluations of LLMs focused on general-purpose knowledge as measured by MMLU \cite{hendrycks2021mmlu}. HELM \cite{liang2023helm} later introduced a unified evaluation framework that aggregates benchmarks across broader dimensions. Although these evaluations characterize overall language understanding, they do not account for the domain-specific requirements or operational constraints encountered in cybersecurity settings.

% Domain-specific cybersecurity benchmark emerged (SecEval, CyberMetric, CyberBench, SECURE, CS-Eval, SecBench, CTI-Bench). They focus on cybersecurity knowledge recall via MCQ or structured-answer questions.
As a result, several domain-specific benchmarks have been proposed to evaluate LLMs in cybersecurity contexts. Benchmarks such as SecEval \cite{li2023seceval}, CyberMetric \cite{cybermetric}, CyberBench \cite{liu2024cyberbench}, SECURE \cite{bhusal2024secure}, CS-Eval \cite{yu2024cseval}, SecBench \cite{secbench}, and CTI-Bench \cite{alam2024ctibench} primarily assess cybersecurity knowledge through multiple-choice, true/false, short-answer, or other classification-style questions spanning software and network security, cryptography, industrial control systems, and cyber-threat intelligence. While these benchmarks provide useful insight into models’ conceptual understanding, they largely evaluate knowledge recall and static reasoning rather than concrete operational behavior.

% More advanced agentic benchmark (CyBench, NYU-CTF)
More recent work moves beyond static question answering toward \emph{agentic} evaluation, where models interact with complex environments or autonomously solve multi-step tasks. For example, CyBench \cite{zhang2024cybench} and NYU-CTF Bench \cite{shao2024nyu-ctf} evaluate LLMs in end-to-end security scenarios via Capture-the-Flag (CTF) challenges, while DefenderBench \cite{zhang2025defenderbench} studies language agents in enterprise-style cybersecurity environments with realistic defensive and offensive workflows. However, end-to-end task performance jointly depends on planning, environment interaction, observation interpretation, tool invocation, and error recovery, making it difficult to attribute failures to individual components. KaliBench complements these benchmarks by isolating command generation and separately evaluating tool selection, optional arguments, and positional arguments.

% However, in these agentic settings, tool usage is embedded within larger workflows, making it difficult to isolate and rigorously evaluate the correctness of individual cybersecurity tool invocations.

\begin{figure*}[!t]
    \centering
    \includegraphics[width=\textwidth]{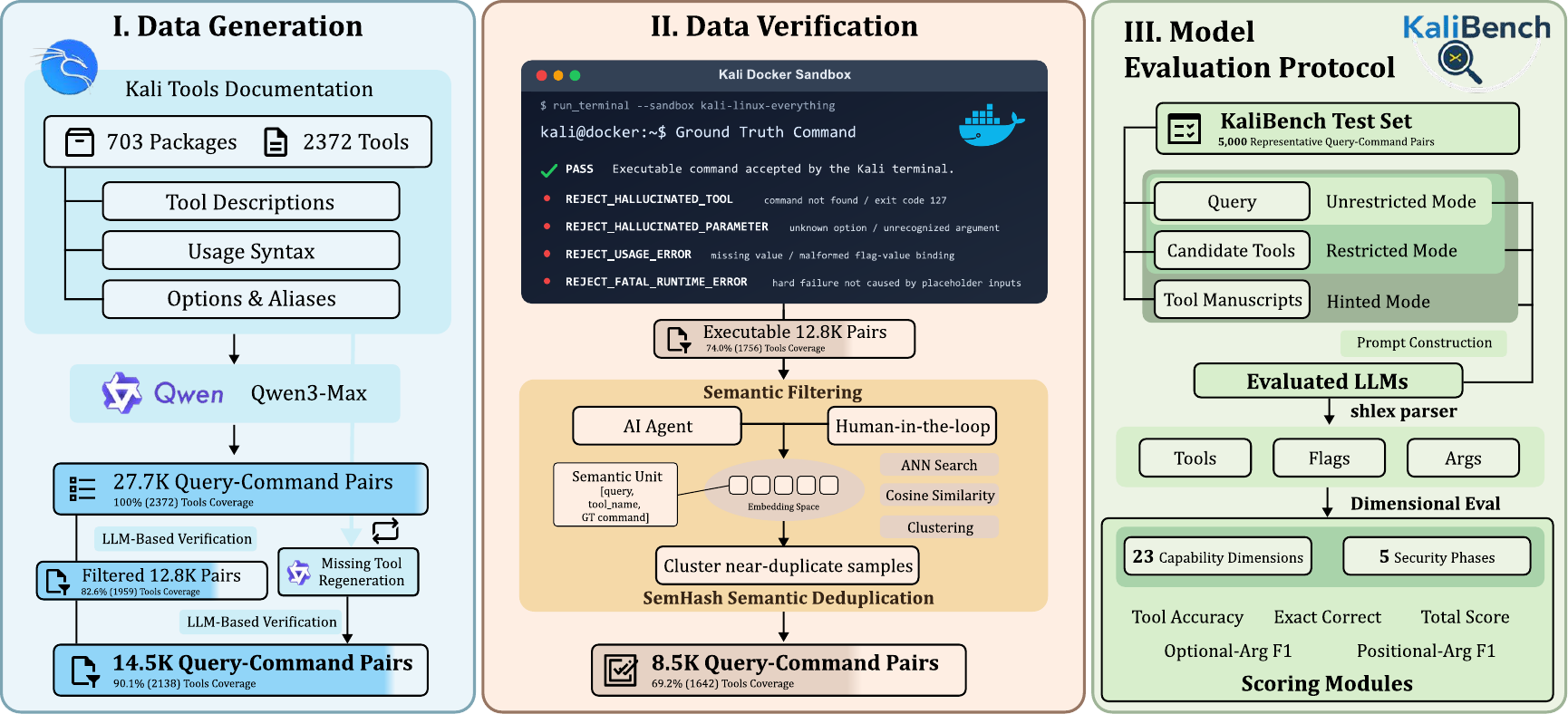}
    \caption{
\textbf{KaliBench end-to-end pipeline for NL-to-CLI evaluation on Kali Linux.}
The benchmark is constructed via LLM-based data generation, multi-round verification-regeneration, and representative-sample selection.
Evaluation applies alias-aware parsing and fine-grained scoring over tools, arguments, and dimensional categories.
}
    \label{fig:kalibench}
    % \vspace{-1em}
\end{figure*}

\paragraph{Function-Calling Benchmarks and Evaluations.}
Beyond agentic evaluation, a growing body of work studies LLMs’ ability to invoke external tools through explicit function-calling benchmarks. Benchmarks such as BFCL \cite{patil2025bfcl}, ToolSandbox \cite{lu2025toolsandbox}, StableToolBench \cite{guo2024stabletoolbench}, and HammerBench \cite{wang2025hammerbench} evaluate tool use across single-turn, multi-turn, conversational, and agentic settings, emphasizing structured, schema-defined APIs and deterministic validation for general-purpose tool invocation. As a result, these benchmarks assume that tools and parameters are specified through explicit schemas, and do not consider settings where tools must be inferred from a shared CLI. To the best of our knowledge, no existing benchmark directly evaluates natural-language–to–CLI translation for real-world cybersecurity tools under such schema-free conditions, or provides fine-grained, alias-aware assessment of command correctness. \textbf{KaliBench} addresses this gap by enabling systematic evaluation of exact, executable CLI command generation for real-world cybersecurity toolchains. 

% \paragraph{Cybersecurity LLM.} ...

% \paragraph{Cybersecurity LLM} \NS{TODO: Optionally add related work on CyberSecurity LLM and how the function-calling model is trained.} \\

\section{KaliBench}

\begin{figure*}[t]
    \centering
\    \includegraphics[width=\textwidth]{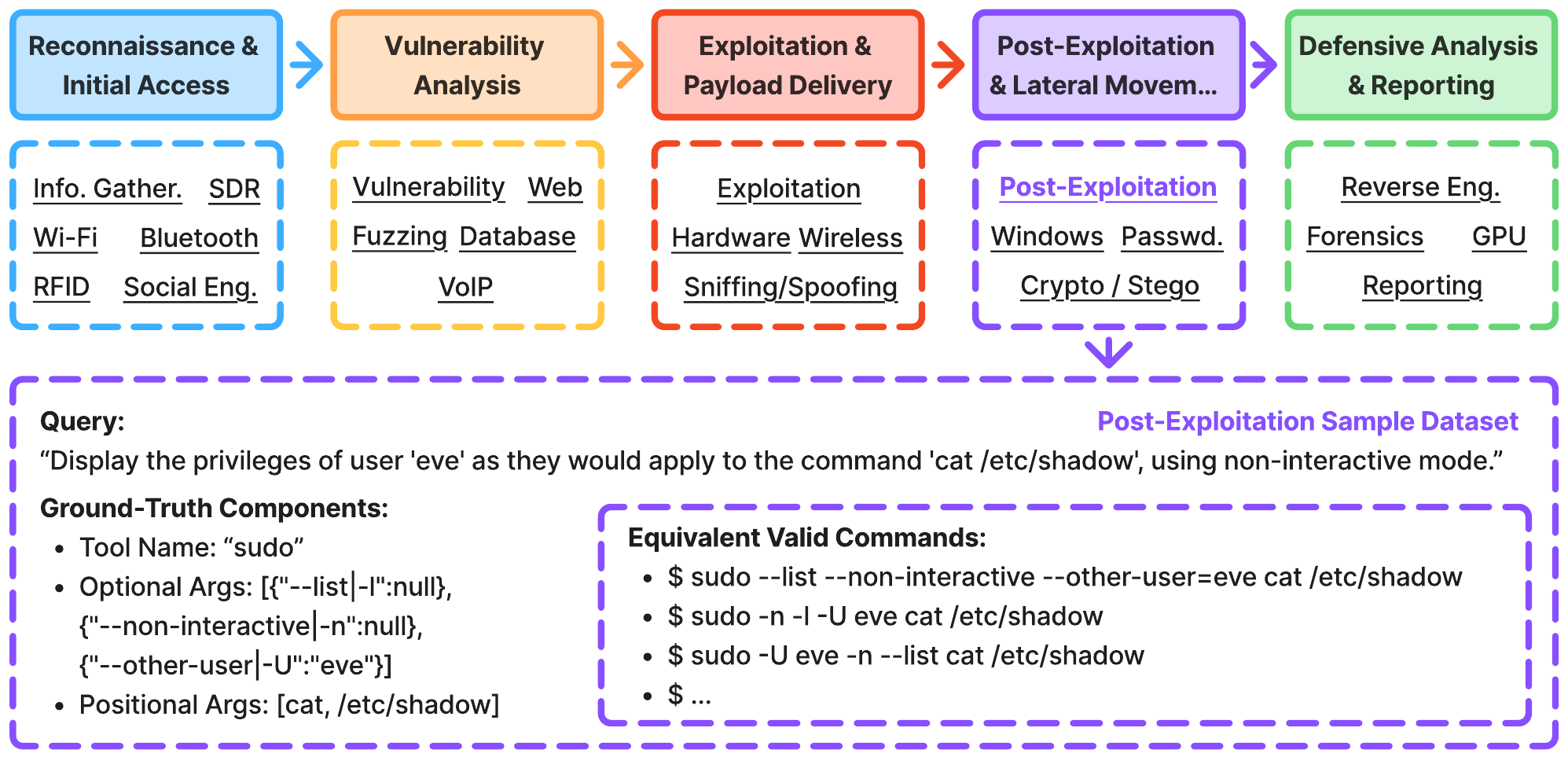}
% \fbox{\rule{0pt}{4.5cm}\rule{0.95\textwidth}{0pt}}
\caption{
\textbf{KaliBench Structure and Representative Sample.}
KaliBench is a structured benchmark of Kali Linux tools organized along the security attack lifecycle.
It comprises 5 security phases and 23 functional dimensions. The dataset sample includes natural-language queries and decomposed ground-truth commands. See Appendix \ref{text} for a detailed view of examples of all 23 dimensions. 
% Zoom in for the best view. 
% \NS{TODO: Refine or remove.}
% \SC{this more like some text in para, not a figure, will try to refine, e.g, give a linux system pipeline and point each sample from which stage ...}
}

    \label{fig:text visualization}
    \vspace{-1mm}
\end{figure*}

% =========== Revised Naufal ============
% spanning \textbf{501 tools}, \textbf{1,662 sub-tools}, and \textbf{11,790} 

KaliBench is a fine-grained benchmark and dataset that serves as a \emph{translation layer} between natural-language cybersecurity queries and executable Kali Linux commands, spanning \textbf{1,642 tools} and \textbf{8,504} query--command pairs. Figure~\ref{fig:kalibench} illustrates the end-to-end pipeline for constructing, verifying, evaluating, and training with KaliBench. The pipeline proceeds in four stages: (1) we extract tool names, flag--value pairs, and documented aliases from official Kali Linux manuscripts, and use them as structured signals to guide LLM-based data generation; (2) generated candidates are iteratively filtered through LLM-based verification, sandboxed terminal execution, and human-in-the-loop evaluation; (3) the verified dataset is deduplicated and split into evaluation and training sets, enabling fine-grained evaluation across multiple metrics and capability dimensions; and (4) using the training split and verifiable reward signals, we demonstrate that parameter efficient post-training with supervised fine-tuning and reinforcement learning significantly improves model performance.

\paragraph{1. Data Generation.}
We curate official documentation for Kali Linux tools from the Kali Linux repository \cite{kalitools2022}.
% \footnote{Kali-Tools Documentation: \url{https://gitlab.com/kalilinux/documentation/kali-tools}}. 
The source corpus comprises 2,372 tools described in structured Markdown, each specifying long and short options and relevant aliases. The documentation also provides a metapackage taxonomy, which we use to annotate each tool with its cybersecurity capability and functional dimension.
We process the corpus at the tool level. For each tool, we use Qwen3-Max \cite{qwen3max_blog_2026} to generate 10-13 natural-language query--command pairs, along with structured annotations including the tool name, optional arguments as alias-aware flag--value pairs, and positional arguments. These ground-truth command components enable fine-grained analysis of model performance in tool selection, argument construction, and syntactic validity under UNIX shell conventions.
Representative examples are shown in Fig.~\ref{fig:text visualization}, illustrating the diversity of KaliBench across security phases and functional dimensions. This stage produces 27.7K query--command pairs covering all documented tools. More details about the data generation process can be found in Appendix \ref{data generation}. 
% \NS{{Finished---- Pengfei: Please add mention of the corresponding appendix for data generation and structure details.}}

\paragraph{2. Data Verification, Cleaning, and Deduplication.}
Despite grounding data generation in official Kali Linux tool manuscripts using a state-of-the-art model (Qwen3-Max), hallucination remains a major challenge. Our initial manual inspection shows that over 50\% of generated commands contain hallucinated or misspelled options (e.g., \verb|--silent| instead of \verb|--silence|). To address this issue, we adopt a \emph{multi-stage human-in-loop verification pipeline} that ultimately retains 8,504 high-quality query--command pairs after cleaning and deduplication. 

\textbf{\textit{LLM-as-Verifier.}} We prompt an LLM with the \texttt{query}, the \texttt{generated\_command}, and the corresponding \texttt{tool\_manuscript}, and task it with verifying (i) whether all flags and arguments are documented, and (ii) whether the command faithfully implements the query using only supported functionality. After the first verification round, 53.8\% of generated examples are filtered out. This leaves 413 tools (17.4\%) without valid pairs; to preserve coverage, we iteratively regenerate and re-verify the data. The resulting LLM-validated dataset contains 14.5K query--command pairs, with 47.65\% of the initial data removed. The prompt details are provided in Appendix \ref{appendix:subsec-llm-verification}.

\textbf{\textit{Terminal Verification.}} To further ensure the executability of ground-truth commands, we perform system-level validation by executing each LLM-verified command in a controlled sandbox. Specifically, we run each command inside a Docker container provisioned with a \emph{Kali-linux-everything} image and record its outputs, including \texttt{stdout}, \texttt{stderr}, exit code, and execution timeouts.
Because command parameters are synthetic (e.g., addresses, devices, paths, and files may not exist in the sandbox), we apply a rule-based filtering pipeline based on observed execution outputs. Commands are rejected when execution logs indicate errors caused by hallucinated tools or invalid arguments that were not detected by the LLM-as-Verifier. This process filters out 11.95\% of the LLM-verified commands. Further details of the terminal verification process are provided in Appendix~\ref{terminal}.

\textbf{\textit{Human-in-the-loop Evaluation.}} As a final verification step, we perform human-in-the-loop evaluation by analyzing model failures in the hinted setting (see the Evaluation Protocol). We focus on cases where even the strongest evaluated model fails to generate a correct command despite having access to tool documentation, as these cases are more likely to reflect issues in the dataset rather than model limitations.
We use an AI agent\footnote{\href{https://github.com/openai/codex}{OpenAI Codex} with \href{https://deploymentsafety.openai.com/gpt-5-4-thinking}{GPT-5.4 Thinking}.} to produce initial verdicts by cross-checking the semantic alignment among the query, generated command, ground-truth command, and tool documentation. Human reviewers then examine these verdicts to identify query ambiguities and labeling errors. This process reveals that an additional 4.9\% of the verified dataset is invalid due to: (a) Malformed CLI options (e.g.,`\dashdash g' instead of the documented `-g'), (b) Omission of arguments explicitly required by the query (e.g., generating `\texttt{<tool> completion}' instead of the query-required `\texttt{<tool> completion bash}'), and (c) Semantic mismatches in argument use (e.g., choosing a flag whose effect does not match the requested operation, swapping the positions of input and output operands, or supplying a filesystem path where the command expects a URL). Detailed cases can be found in Appendix \ref{manual case}.

% \NS{Pengfei: please summarize your findings during manual verification. Examples: errors in queries (e.g., ambiguity, missing constraints or context); errors in ground-truth commands (e.g., incorrect argument parsing, unsupported options, or misclassified positional arguments).}. 

\textbf{\textit{Deduplication and Train--test Split.}} To construct a compact yet semantically diverse evaluation set, we perform \emph{semantic selection} using the SemHash framework \cite{minishlab2025semhash}. Each record is represented as a semantic unit [\texttt{query}, \texttt{tool\_name}, \texttt{ground\_truth\_command}]. Records are embedded into a dense vector space, clustered via approximate nearest-neighbor search, and grouped when pairwise cosine similarity exceeds 0.90. From each cluster, we retain representative examples selected by SemHash.
The final cleaned and deduplicated dataset contains \textbf{8.5K verified query--command pairs} (30.7\% of the original 27.7K generated data), forming a high-quality and executable benchmark for NL--to--CLI evaluation. We then sample from this dataset to construct an evaluation set of 5,000 examples (covering all 1,642 tools) and a training set of 3,504 examples (covering 962 tools), prioritizing full tool coverage in the evaluation set while reserving a smaller training split for studying generalization.

% \begin{figure}[t]
% \centering
% \scriptsize
% \fbox{
% \begin{tabular}{p{0.94\linewidth}}
% \textbf{Query:} \\
% \textit{Display the privileges of user 'eve' as they would apply to the command 'cat /etc/shadow', using non-interactive mode.} \\[4pt]

% \textbf{Ground-truth components:} \\
% \texttt{- tool\_name: sudo} \\
% \texttt{- optional\_args: [\{"\dashdash list|-l":null\}, \{"\dashdash non-interactive|-n":null\}, \{"\dashdash other-user|-U":"eve"\}]} \\
% \texttt{- positional\_args: [cat, /etc/shadow]} \\[4pt]

% \textbf{Equivalent valid commands:} \\
% \texttt{\$ sudo \dashdash list \dashdash non-interactive \dashdash other-user=eve cat /etc/shadow} \\
% \texttt{\$ sudo -n -l -U eve cat /etc/shadow} \\
% \texttt{\$ sudo -U eve -n \dashdash list cat /etc/shadow}
% \end{tabular}
% }
% \caption{KaliBench example showing format, component breakdown, and command equivalence.}
% \vspace{-1em}
% % \caption{Example from KaliBench illustrating the dataset format, ground-truth component decomposition, and alias-aware command equivalence.}
% \label{fig:cmd_parsing_example}
% \end{figure}

\paragraph{3. Evaluation Protocol.}
For each query in the evaluation set, models are prompted to generate an executable Kali Linux CLI command under three settings that vary the amount of tool information provided: (1) \emph{Unrestricted}, where the model receives only the natural-language query and must infer both the appropriate tool and its arguments; (2) \emph{Restricted}, where the model is additionally provided with a randomly sampled list of 20 candidate tool names that includes the correct tool, reducing the tool-selection search space while still requiring correct argument construction; and (3) \emph{Hinted}, where the model further receives the official documentation of the candidate tools, approximating schema-based function-calling benchmarks in which tool functionality and parameters are explicitly specified. Details of the structured prompting format and prompt templates are provided in Appendix~\ref{evaluation_prompt}.

\textbf{\textit{Command Parsing.}}
Model outputs are parsed using Python’s \texttt{shlex} library for lexical analysis of shell-style syntax, enabling faithful emulation of Kali Linux command parsing under UNIX conventions without executing commands. Because KaliBench explicitly records the ground-truth \texttt{tool\_name}, optional arguments (as alias-aware flag--value pairs), and ordered positional arguments for each example, we can perform deterministic, component-wise scoring.
A CLI command follows the general form \texttt{<tool\_name> <args>}, where arguments are either optional or positional. Optional arguments begin with \verb|--| or \verb|-| and may take values; their order is flexible and aliases are permitted. Positional arguments consist only of values and must appear in a fixed order. See Fig.~\ref{fig:text visualization} for illustration.
% Fig.~\ref{fig:text visualization} illustrates the KaliBench data format and the corresponding ground-truth decomposition.

\textbf{\textit{Evaluation Metrics.}}
We report a set of fine-grained, component-level metrics, including \textit{Tool Accuracy} (correct tool selection), \textit{Optional-Argument F1} (alias-aware matching of optional flag names), \textit{Positional-Argument F1} (multiset matching of positional argument values), \textit{Total Score} (mean of tool and argument scores), and \textit{Exact Correct}, which equals 1 only when the tool and all arguments exactly match the ground truth. These metrics enable detailed analysis of model performance across tool selection and argument construction. Details of the metric definitions and calculations are provided in Appendix~\ref{metric_cal}.

\textbf{\textit{Capability Taxonomy and Security-Phase Annotation.}}\label{kali_tool_dim23}
To enable structured, domain-aware analysis beyond raw command correctness, we leverage capability metadata derived from Kali Linux’s official \texttt{kali-tools} metapackage taxonomy, which organizes offensive and defensive tools into 23 functional capability dimensions spanning the full cybersecurity workflow. Inspired by the Lockheed Martin Cyber Kill Chain~\cite{hutchins2011killchain}, we group these dimensions into five security phases: \emph{Reconnaissance \& Initial Access} (information-gathering, 802-11, rfid, bluetooth, sdr, social-engineering), \emph{Vulnerability Analysis} (vulnerability, fuzzing, web, database, voip), \emph{Exploitation \& Payload Delivery} (exploitation, sniffing-spoofing, hardware, wireless), \emph{Post-Exploitation \& Lateral Movement} (post-exploitation, passwords, windows-resources, crypto-stego), and \emph{Defensive Analysis \& Reporting} (reverse-engineering, forensics, reporting, gpu). Although some capabilities span multiple stages, each dimension is assigned a primary functional role for clarity, enabling aggregation of evaluation metrics across coherent capability dimensions and security phases and supporting systematic analysis of model performance in natural-language--to--CLI translation across different cybersecurity contexts.

\paragraph{4. Training Protocol.}
\label{training-protocol}
Our training data support LLM post-training via Supervised Fine-Tuning (SFT) and Reinforcement Learning with Verifiable Rewards (RLVR). We adopt the same prompt template used during evaluation, with the query, a randomly sampled list of candidate tools, and tool documentation as inputs. We concatenate data from all three evaluation settings, resulting in $3{,}504 \times 3 = 10{,}512$ training samples. Additional details are provided in Appendix~\ref{appendix:training-protocol}.

\begin{wrapfigure}{r}{0.42\linewidth}
    \centering
    \vspace{-10pt}
    \includegraphics[width=\linewidth]{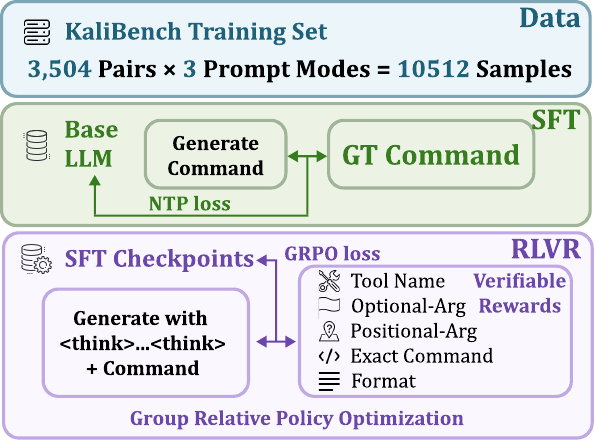}
    \caption{
    KaliBench Training Protocol.
    % \SC{plz add caption}
    % \NS{How about change "REDSAGE INS-8B" with just "LLM"?}\SC{ok}
    }
    \label{fig:train_pipe}
    \vspace{-12pt}
\end{wrapfigure}

\textbf{\textit{SFT.}} We use the constructed prompt template and treat the ground-truth command as the target output. The model is trained to generate the correct command given the query and optional tool information.

\textbf{\textit{RLVR.}} We extend the prompt template by instructing the model to produce intermediate reasoning within \emph{<think>...</think>} tags before generating the final command, following \cite{deepseek-r1}. Our reward function is derived from component-level evaluation metrics, providing fully verifiable signals for accurate prediction of tool name, positional arguments, and alias-aware optional arguments. This enables fine-grained learning signals beyond a binary exact-match objective. An additional reward is assigned for exact command matches. Because ground-truth commands are validated during data construction via sandboxed execution, rewards can be computed without executing model outputs, enabling efficient RL without runtime. We include a formatting reward to enforce output structure.

%%%%%%%%%%%%%%%%%%%%%%%%%%% New 5000 verified eval sets %%%%%%%%%%%
\begin{table*}[!t]
\centering
\caption{\textbf{Open-weight model query-to-command performance across three modes.} Columns report Unrestricted (U), Restricted (R), and Hinted (H). $^{\dagger}$ indicates thinking-enabled models, and $^{\star}$ denotes cybersecurity-tuned models. \textbf{Bold}/\underline{underline} mark the best and second-best within each size group. Models are sorted by Averaged Total Score. MoE parameters use Total/Active (e.g., 685/37B).}\label{tab:combined_perf}

\resizebox{\linewidth}{!}{%
\begin{tabular}{l | c |
    c c c |
    c c c
    c c c
    c c c | 
    c c c  c
}
\toprule
% \rowcolor{blue!10}
\multirow{2}{*}{\textbf{Model}} & \textbf{Size} &
\multicolumn{3}{c|}{\textbf{Exact Correct (\%)}} &
\multicolumn{3}{c}{\textbf{Tool Acc. (\%)}} &
\multicolumn{3}{c}{\textbf{Optional F1 (\%)}} &
\multicolumn{3}{c}{\textbf{Positional F1 (\%)}} &
\multicolumn{4}{|c}{\textbf{Total Score (\%)} } 
% &
% \textbf{Avg.$^\diamond$}
\\
\cmidrule(lr){3-5} \cmidrule(lr){6-8} \cmidrule(lr){9-11} \cmidrule(lr){12-14} \cmidrule(lr){15-18}
 & (B) & U & R & H & U & R & H & U & R & H & U & R & H & U & R & H & Avg. \\
\midrule

\multicolumn{18}{>{\columncolor{gray!20}}c}{\textbf{Compact Models ($\leq$ 8B)}} \\
\midrule

Foundation-Sec-Ins$^{\star}$ & 8
& 14.7 & 19.1 & 56.2 & 66.5 & 92.5 & 86.8 & 33.5 & 37.4 & 75.6 & 53.0 & 60.0 & 81.3 & 51.0 & 63.3 & 81.2 & 65.2 \\

Llama3.1-Ins & 8
& 12.5 & 18.2 & 62.7 & 61.7 & 93.7 & 95.4 & 32.4 & 37.2 & 82.7 & 53.5 & 61.9 & 85.8 & 49.2 & 64.3 & 88.0 & 67.2 \\

Llama-Primus-Base$^{\star}$ & 8
& 15.8 & 19.3 & 60.3 & 67.3 & 90.7 & 94.1 & 36.6 & 39.2 & 80.8 & 56.3 & 61.9 & 84.0 & 53.4 & 63.9 & 86.3 & 67.9 \\

Foundation-Sec-Rsn$^{\star}$$^{\dagger}$ & 8
& 13.3 & 18.6 & 66.6 & 67.0 & 92.8 & 95.9 & 34.8 & 39.1 & 85.5 & 53.0 & 57.9 & 86.2 & 51.6 & 63.3 & 89.2 & 68.0 \\

Llama-Primus-Merged$^{\star}$ & 8
& 16.3 & 21.0 & 65.2 & 66.2 & 93.4 & 95.2 & 36.7 & 40.8 & 84.1 & 56.6 & 63.2 & 87.1 & 53.2 & 65.8 & 88.8 & 69.3 \\

RedSage-DPO$^{\star}$ & 8
& 18.2 & 25.4 & 68.1 & 66.8 & 95.1 & 95.7 & 40.1 & 46.1 & 85.4 & 56.7 & 65.9 & 88.5 & 54.5 & 69.0 & 89.9 & 71.1 \\

Qwen3  & 8
& 18.3 & 24.1 & 73.1 & 68.9 & 95.6 & \underline{96.3} & 39.3 & 43.4 & 88.2 & 58.5 & 65.6 & 90.5 & 55.5 & 68.2 & 91.6 & 71.8 \\

Qwen3$^{\dagger}$ & 8
& 20.9 & 27.7 & \textbf{80.9} & 70.3 & \underline{95.7} & 96.2 & 42.9 & 47.2 & \textbf{92.4} & 59.9 & 66.5 & \textbf{93.4} & 57.7 & 69.8 & \textbf{94.0} & 73.8 \\

\midrule

RedSage-Ins$^{\star}$ & 8
& 19.7 & 26.1 & 67.2 & 67.9 & 94.6 & 95.3 & 42.4 & 46.9 & 85.0 & 58.8 & 66.4 & 88.3 & 56.4 & 69.3 & 89.5 & 71.7 \\

\rowcolor{blue!10}
\textbf{RedSage-K (GRPO)$^{\star}$$^{\dagger}$} & 8
& 28.1 & 33.0 & 66.7 & 76.5 & 93.3 & 93.1 & 51.7 & 53.7 & 84.8 & \underline{73.9} & \underline{77.0} & 88.4 & \underline{67.4} & \underline{74.6} & 88.8 & 76.9 \\

\rowcolor{blue!10}
\textbf{RedSage-K (SFT)$^{\star}$ }& 8
& \underline{30.6} & \underline{34.4} & \underline{77.1} & \textbf{78.1} & \textbf{95.9} & \textbf{97.1} & \underline{52.8} & \underline{55.2} & \underline{90.5} & 66.9 & 68.4 & \underline{92.0} & 65.9 & 73.2 & \underline{93.2} & \underline{77.4} \\

\rowcolor{blue!10}
\textbf{RedSage-K (SFT+GRPO)$^{\star}$$^{\dagger}$} & 8
& \textbf{32.2} & \textbf{37.4} & 69.4 & \underline{77.9} & 95.0 & 92.5 & \textbf{56.1} & \textbf{57.9} & 86.5 & \textbf{76.9} & \textbf{80.9} & 89.1 & \textbf{70.3} & \textbf{77.9} & 89.3 & \textbf{79.2} \\

\midrule
\multicolumn{18}{>{\columncolor{gray!20}}c}{\textbf{Mid-size Models (20-32B)}} \\
\midrule

Gemma-3-it & 27
& 18.9 & 24.0 & 70.5 & \underline{73.4} & \textbf{97.3} & 87.0 & 40.9 & 44.2 & 83.1 & 60.2 & 66.3 & 88.4 & 58.1 & 69.2 & 86.2 & 71.2 \\

GPT-OSS$^{\dagger}$ & 20/3.6
& 19.6 & 24.6 & 78.1 & 72.3 & 95.1 & \underline{95.9} & 42.2 & 45.6 & 91.3 & 57.1 & 63.6 & 92.0 & 57.2 & 68.1 & 93.0 & 72.8 \\

Mistral-Small-3.2 & 24
& \underline{20.7} & 27.4 & \underline{79.9} & 70.6 & 95.6 & \textbf{96.2} & \underline{44.3} & 48.2 & \underline{91.5} & 62.6 & 69.2 & 93.2 & 59.2 & 71.0 & \underline{93.7} & 74.6 \\

Qwen3 & 32
& 20.6 & \underline{27.7} & 79.8 & \textbf{73.5} & 96.3 & 95.5 & 43.4 & \underline{48.9} & 91.2 & \underline{62.8} & \underline{69.9} & \underline{93.4} & \underline{59.9} & \underline{71.7} & 93.4 & \underline{75.0} \\

Qwen3$^{\dagger}$ & 32
& \textbf{24.0} & \textbf{30.9} & \textbf{81.8} & 72.4 & \underline{96.8} & 95.2 & \textbf{47.2} & \textbf{51.6} & \textbf{92.4} & \textbf{65.3} & \textbf{71.4} & \textbf{94.0} & \textbf{61.6} & \textbf{73.3} & \textbf{93.9} & \textbf{76.3} \\

\midrule
\multicolumn{18}{>{\columncolor{gray!20}}c}{\textbf{Large-scale Models ($\geq$ 70B)}} \\
\midrule

Llama-Primus-Ins$^{\star}$ & 70
& 18.8 & 25.0 & 71.2 & 72.8 & 96.4 & 92.1 & 41.9 & 46.3 & 88.1 & 59.6 & 67.0 & 90.2 & 58.1 & 69.9 & 90.1 & 72.7 \\

Llama-3.3-Ins & 70
& 22.6 & 28.7 & 77.7 & 72.1 & 96.5 & 94.1 & 45.7 & 49.9 & 91.7 & 61.9 & 68.1 & 92.5 & 59.9 & 71.5 & 92.7 & 74.7 \\

GPT-OSS$^{\dagger}$ & 120/5.1
& 24.0 & 32.0 & \underline{80.9} & 73.6 & 95.6 & 96.0 & 48.0 & 52.9 & \underline{92.2} & 61.4 & 68.7 & 92.7 & 61.0 & 72.4 & 93.6 & 75.7 \\

Qwen3-Coder-Next$^{\dagger}$ & 80/3
& 26.2 & 31.0 & 79.9 & 74.1 & 96.8 & 96.4 & 48.8 & 51.9 & 91.4 & 63.9 & 70.7 & 93.2 & 62.3 & 73.1 & 93.6 & 76.3 \\

Qwen2.5-Ins & 72
& 25.4 & 30.2 & 80.6 & \underline{76.0} & \underline{97.4} & \underline{96.8} & 47.7 & 51.1 & 91.5 & \underline{67.0} & 72.2 & \underline{93.3} & 63.6 & 73.6 & \underline{93.8} & 77.0 \\

DeepSeek-V3.2$^{\dagger}$ & 685/37
& \underline{33.0} & \underline{40.2} & \textbf{84.5} & 75.8 & \textbf{98.3} & \textbf{97.1} & \underline{57.0} & \underline{62.0} & \textbf{93.5} & \textbf{68.8} & \underline{75.4} & \textbf{94.1} & \underline{67.2} & \underline{78.6} & \textbf{94.9} & \underline{80.2} \\ 

GLM-5.2$^{\dagger}$ & 753
& \textbf{41.3} & \textbf{52.0} & 74.8 & \textbf{86.3} & 94.2 & 94.9 & \textbf{75.3} & \textbf{71.9} & 88.4 & 51.3 & \textbf{78.8} & 90.6 & \textbf{71.0} & \textbf{81.7} & 91.3 & \textbf{81.3} \\ 

\midrule
\rowcolor{gray!10}
\textbf{Metric Average} & --
  & 22.3 & 28.3 & 73.1
  & 72.0 & 95.2 & 94.6
  & 45.1 & 48.7 & 87.8
  & 61.1 & 68.2 & 90.1
  & 59.4 & 70.7 & 90.8
  & 73.6 \\

\bottomrule
\end{tabular}%
 }
 % \vspace{-1mm}
\end{table*}

% & 21.6 & 27.4 & 73.0
% & 71.4 & 95.2 & 94.6
% & 43.9 & 47.8 & 87.8
% & 61.6 & 67.8 & 90.1
% & 59.0 & 70.3 & 90.8
% & 73.4 \\

\section{Experiments}

\paragraph{Experiment Setup.}
All evaluations of open-source models were conducted on an NVIDIA A100 multi-GPU cluster using the vLLM inference engine, with selected large models evaluated via external APIs\footnote{OpenRouter: \url{https://openrouter.ai/}}.
We evaluate 14 general-purpose models: \textit{Llama3.1-Instruct (8B)}~\cite{llama3}, \textit{Qwen-3 (8B, 32B; with and without reasoning)}~\cite{yang2025qwen3}, \textit{Mistral-3.2 (24B)}~\cite{mistral_small_3_2_24b_2025}, \textit{Gemma-3 (27B)}~\cite{gemma3}, \textit{GPT-OSS (20B, 120B)}~\cite{gptoss-20-120}, \textit{LLaMA-3.3-Instruct (70B)}~\cite{llama3_3_70b_instruct_2024}, \textit{Qwen-2.5-Instruct (72B)}~\cite{qwen2025qwen25technicalreport}, \textit{Qwen3-Coder-Next (80B)}~\cite{qwen3codernext}, and \textit{DeepSeek-V3.2 (685B)}~\cite{deepseekv32}, \textit{GLM-5.2 (753B)}~\cite{glm5.2} along with 7 cybersecurity-oriented models: \textit{Foundation-Sec-Instruct (8B)}~\cite{weerawardhena2025sec8b}, \textit{Foundation-Sec-Reasoning (8B)}~\cite{funsec-reasoning}, \textit{Llama-Primus-Base/Merged (8B)} and \textit{Llama-Primus-Nemotron (70B)}~\cite{yu2025primuspioneeringcollectionopensource}, and \textit{RedSage-Ins/DPO (8B)}~\cite{suryanto2026redsage}, as well as three variants of a cybersecurity-oriented model further fine-tuned on KaliBench.
Models with 72B or more parameters were evaluated through the OpenRouter API; all other models are obtained from their official Hugging Face repositories under their respective licenses and evaluated under a fixed inference configuration; details are in Appendix~\ref{appendix:evaluated-model}.

\paragraph{Training Setup.}
We employ parameter-efficient fine-tuning using LoRA \cite{hu2022lora} with Unsloth \cite{unsloth}. We select the best-performing 8B cybersecurity-tuned model \textit{RedSage-Ins} as the backbone. Using the training dataset described in Sec.~\ref{training-protocol}, we train three variants, collectively referred to as \textit{RedSage-K}: (1) SFT, (2) RLVR with Group Relative Policy Optimization (GRPO) \cite{shao2024deepseekmath}, and (3) SFT followed by GRPO (SFT+GRPO). All models are trained on a single H200 GPU (141GB VRAM) with a total batch size of 32 for two epochs; details are in Appendix~\ref{appendix:training-protocol}.

\subsection{Model Performance Across Three Modes}

% Table~\ref{tab:combined_perf} summarizes per-model performance under three evaluation modes: \textit{Unrestricted (U)}, \textit{Restricted (R)}, and \textit{Hinted (H)}. We report Tool Accuracy, Optional-Argument F1, Positional-Argument F1, Total Score, and Exact Correct for each model and mode. We analyze key patterns in these results, focusing on the impact of tool hints, model type and scale, and post-training on KaliBench.

Table~\ref{tab:combined_perf} summarizes per-model performance under three modes: \textit{Unrestricted (U)}, \textit{Restricted (R)}, and \textit{Hinted (H)}. We report Exact Correct (EC), Tool Accuracy (TA), Optional-Argument F1 (Opt-F1), Positional-Argument F1 (Pos-F1), Total Score (TS), and Average TS (Avg.). We analyze key patterns with respect to tool information, model type and scale, and post-training on KaliBench.

\paragraph{Impact of Evaluation Modes.}
The three evaluation modes reveal a clear separation between tool selection and argument construction. Transitioning from \textit{U} to \textit{R} substantially improves TA (72.0\%$\rightarrow$95.2\% on average), indicating that constraining the candidate tool space largely resolves tool-selection errors. However, this shift yields only modest gains in Opt-F1, Pos-F1, and EC, indicating that correct tool selection alone does not ensure accurate argument generation. The metric averages further reveal a hierarchy of difficulty: optional-argument prediction is the most challenging, followed by positional-argument prediction.
In contrast, \textit{H} produces the largest improvements, increasing Opt-F1 from 45.1\% to 87.8\%, Pos-F1 from 61.1\% to 90.1\%, TS from 59.4\% to 90.8\%, and EC from 22.3\% to 73.1\%. These gains indicate that tool hints resolve ambiguities in argument selection, ordering, and formatting, and suggest that existing models lack sufficient domain knowledge to construct accurate commands without tool documentation.
Overall, \textit{R} primarily improves \emph{which} tool to use, whereas \textit{H} improves \emph{how} the command is constructed; without hints, open-weight models still struggle with accurate argument generation, the primary bottleneck.

\paragraph{Impact of Model Type and Scale.}
Cybersecurity-oriented models ($^\star$) do not consistently outperform strong general-purpose models on query-to-command generation, but their advantages become more apparent in less-guided settings. In the compact regime, \textit{RedSage-Ins$^{\star}$} achieves higher TS than non-thinking \textit{Qwen3} in both \textit{U} and \textit{R} (56.4\% vs. 55.5\% and 69.3\% vs. 68.2\%), suggesting that domain-specific training provides useful priors when models must infer command structure with limited support. In contrast, general-purpose models perform better in \textit{H}, where \textit{Qwen3} surpasses \textit{RedSage-Ins$^{\star}$} (91.6\% vs. 89.5\%), indicating stronger instruction-following and context utilization under richer guidance. Overall, cybersecurity-oriented models are more effective in under-specified settings, while general-purpose models better exploit explicit hints.
Reasoning-enabled variants ($^{\dagger}$) further improve argument-level performance, particularly in Opt-F1, Pos-F1, and EC. For example, \textit{Qwen3$^{\dagger}$} improves Avg from 71.8\% to 73.8\% over \textit{Qwen3} and achieves strong \textit{H}-mode results (94.0\% TS, 80.9\% EC), highlighting the role of reasoning in intent decomposition and context utilization.
Finally, while performance generally improves with scale, gains are uneven. For example, \textit{Qwen3$^{\dagger}$} (8B) outperforms larger models, including \textit{Gemma-3-it} (27B) and \textit{Llama-Primus-Ins$^{\star}$} (70B), in Avg. These results suggest that effective NL--to--CLI generation depends on domain knowledge, instruction-following, reasoning, and task alignment, rather than scale alone.

\paragraph{Proprietary Models and Scaffolded Inference.}
Our main evaluation emphasizes open-weight models because local deployment is particularly important in cybersecurity, where queries, target information, and operational data may be sensitive and cannot always be shared with proprietary APIs. To assess whether KaliBench remains challenging for stronger proprietary models and scaffolded systems, we additionally evaluate GPT-5.6-Sol~\cite{openai_gpt56sol}, Claude Opus~5~\cite{anthropic_opus5}, and OpenAI Codex ~\cite{openai_codex_cli}, powered by GPT-5.5~\cite{openai_gpt55} with \texttt{xhigh} reasoning effort. We evaluate all systems on the 5,000 test queries in the \textit{Unrestricted} setting.
%We report both answered-only and full-set Exact Correct (EC) which counts blocked or missing outputs as incorrect.

\begin{wraptable}{r}{0.43\linewidth}
      \centering
      \vspace{-10pt}
      \caption{Proprietary models and scaffolded inference in Unrestricted mode.}
      \label{tab:proprietary_unrestricted}

      \small
      \setlength{\tabcolsep}{3pt}
      \renewcommand{\arraystretch}{1.08}
      \begin{tabular}{@{}lrrr@{}}
          \toprule
          \textbf{Model}
          & \makecell{\textbf{Answered}\\\textbf{queries}}
          & \makecell{\textbf{EC (\%)}}
          & \makecell{\textbf{EC (\%)}\\\textbf{(all 5K)}} \\
          \midrule
          GPT-5.6-Sol
          & 4,988 & 61.83 & \textbf{61.68} \\
          Claude Opus 5
          & 3,675 & 59.89 & 44.02 \\
          Codex (GPT-5.5)
          & 4,633 & 55.77 & 51.68 \\
          \bottomrule
      \end{tabular}

      \vspace{3pt}
      \parbox{\linewidth}{\footnotesize
        EC (Exact Correct) is computed over answered queries; the EC (all 5K) counts blocked or missing outputs as incorrect. For Codex, we use its scaffolded CLI workflow and score only the final submitted command.
      }
      \vspace{-10pt}
  \end{wraptable}

As shown in Table~\ref{tab:proprietary_unrestricted}, GPT-5.6-Sol achieves the highest full-set Exact Correct of 61.68\%, followed by Codex at 51.68\% and Claude Opus~5 at 44.02\%. All three outperform GLM-5.2, the strongest evaluated open-weight model in this setting, which achieves 41.32\%. Response coverage, however, varies across systems: GPT-5.6-Sol answers 4,988 of 5,000 queries, compared with 4,633 for Codex and 3,675 for Claude Opus~5. 
These coverage differences suggest that provider-specific safety restrictions influence cybersecurity-related usage. For example, Claude Opus~5 achieves 59.89\% EC on answered queries but answers only 73.50\% of the test set, reducing its full-set EC to 44.02\%. Moreover, despite the gains over open-weight models, the highest full-set accuracy among these systems is 61.68\%, indicating that KaliBench remains unsaturated and that exact command generation is challenging. Complete metrics and reporting details are provided in Appendix~\ref{app:propri}.

% These results show the influence of provider-specific safety restrictions on cybersecurity-related usage. For example, Claude Opus~5 achieves 59.89\% Exact Correct on answered queries, but the provider's
% security policy blocks 26.50\% of queries, reducing its full-set accuracy to 44.02\%. 
\paragraph{Impact of KaliBench Post-training.}
Post-training on the KaliBench training split substantially improves compact models. The three \textit{RedSage-K} variants achieve Avg scores of 76.9, 77.4, and 79.2, with \textit{RedSage-K (SFT+GRPO)} ranking third in Avg among all evaluated open-weight models, only 1.0 point below \textit{DeepSeek-V3.2$^{\dagger}$} despite using an 8B backbone.
The methods exhibit complementary strengths: \textit{RedSage-K (SFT)} attains the best compact-model TA across all modes, shows stronger gains in Opt-F1, and delivers the largest improvements in \textit{H}, suggesting that supervised fine-tuning effectively learns tool mapping and benefits from contextual hints. In contrast, \textit{RedSage-K (GRPO)} shows stronger gains in Pos-F1 and TS under \textit{U} and \textit{R}, but is weaker in \textit{H}.

Combining both methods, \textit{RedSage-K (SFT+GRPO)} yields the best compact-model \textit{U}/\textit{R} TS and EC, demonstrating the benefit of combining SFT and RLVR under limited guidance. However, \textit{RedSage-K (GRPO)} and \textit{RedSage-K (SFT+GRPO)} slightly reduce \textit{H-mode} TS relative to the original \textit{RedSage-Ins} model. Our paired analysis identifies argument-construction errors and regressions on previously solved examples, with SFT initialization mitigating these regressions (Appendix~\ref{app:rlvr_failure_analysis}). Overall, KaliBench post-training enables compact models to approach larger ones in cybersecurity tool use under less-guided settings.

% \paragraph{Impact of KaliBench Post-training.}
% Post-training on the KaliBench training split substantially improves compact models. The three KaliBench-trained 8B variants achieve Avg scores of 76.9, 77.4, and 79.2, with \textit{Our SFT+GRPO} ranking third in average TS among all evaluated open-weight models, only 1.0 point below \textit{DeepSeek-V3.2$^{\dagger}$} despite using an 8B backbone. 

% The methods exhibit complementary strengths: \textit{Kali-SFT} attains the best compact-model TA across all modes, shows stronger gains in Opt-F1, and delivers the largest improvements in \textit{H}, suggesting that supervised fine-tuning effectively learns tool mapping and benefits from contextual hints. In contrast, \textit{Kali-GRPO} shows stronger gains in Pos-F1 and TS under \textit{U} and \textit{R}, but is weaker in \textit{H}. 

% Combining both, \textit{SFT+GRPO} yields the best compact-model \textit{U}/\textit{R} TS and EC, demonstrating the benefit of combining SFT and RLVR under limited guidance. However, GRPO and SFT+GRPO slightly reduce \textit{H-mode} TS relative to the original RedSage-Ins model. Our paired analysis identifies argument-construction errors and regressions on previously solved examples, with SFT initialization mitigating these regressions (Appendix~\ref{app:rlvr_failure_analysis}). Overall, KaliBench post-training enables compact models to approach larger ones in cybersecurity tool use under less-guided settings.

\begin{figure*}[t]
    \centering
    \includegraphics[width=\textwidth]{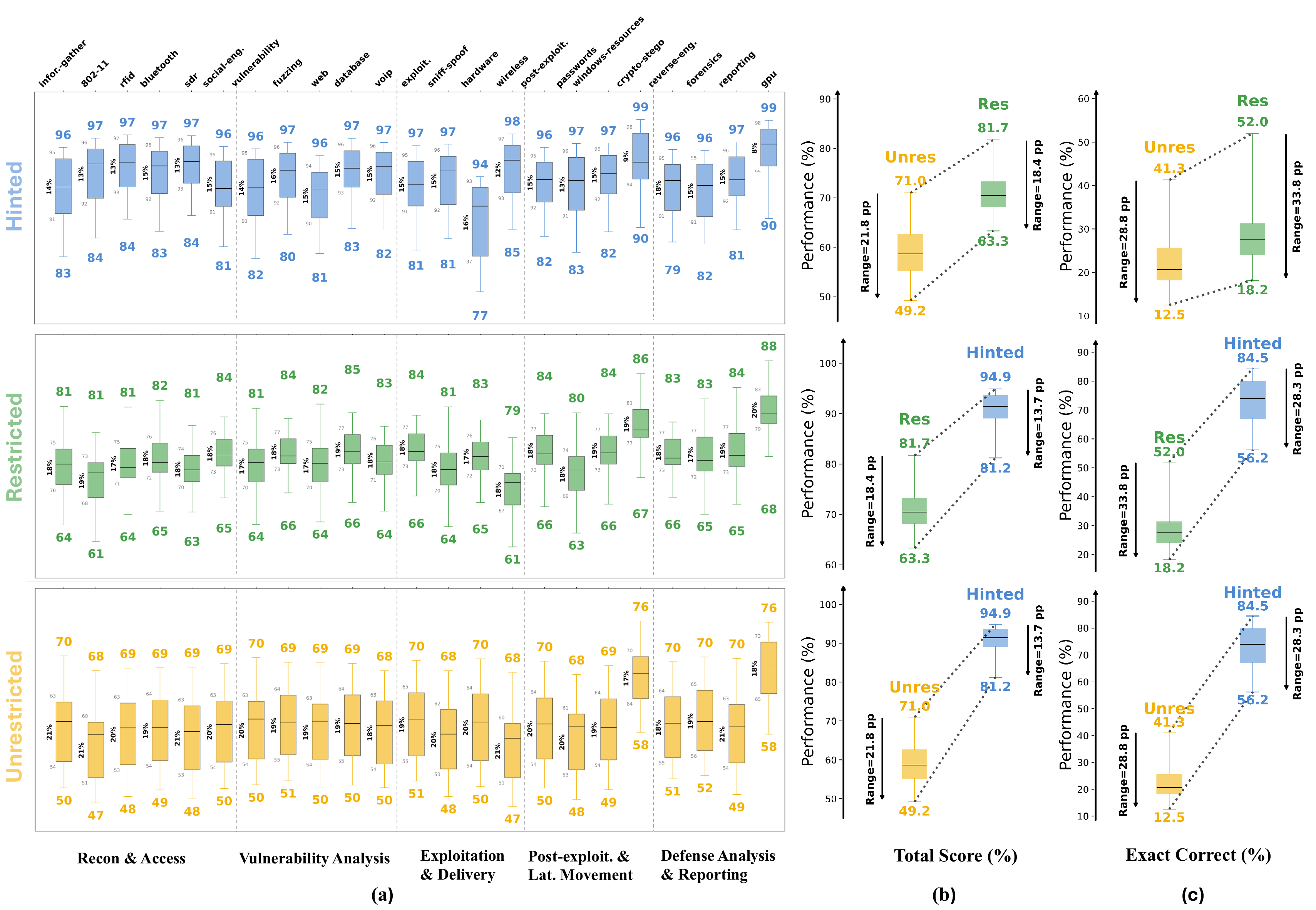}

\caption{
\textbf{Performance across security dimensions and modes.}
(a) Performance over 23 dimensions and 5 phases per mode. Box plots show distributions (Q1--Q3), with whiskers for min/max.
(b,c) Total Score and Exact Correct (\%) across modes; dotted lines indicate inter-mode gaps.
}
    \label{fig:dim}
    \vspace{-1em}
\end{figure*}

% \subsection{Analysis Across Security Dimensions, Performance Distribution, and Failure Cases}

\subsection{Fine-Grained Performance and Robustness Analysis}
\paragraph{Dimension-wise and Tool-level Analysis.}
Figure~\ref{fig:dim}(a) reports TS performance across 23 capability dimensions, five security phases, and three evaluation modes, aggregated over all evaluated models. Performance varies across dimensions, but relative difficulty remains broadly consistent across modes, suggesting that additional tool information improves overall scores without fundamentally changing which dimensions are harder. Dimensions such as \textit{gpu} and \textit{crypto-stego} consistently achieve higher scores, whereas \textit{802-11} and \textit{wireless} show lower minima, especially in \textit{U} and \textit{R} modes. We also observe mode-specific deviations, such as \textit{hardware} in \textit{H} mode, where performance remains comparatively lower, likely reflecting the difficulty of longer or more complex tool descriptions.

From a distributional perspective, variability decreases as tool information increases. The \textit{U} mode shows the widest spread across models, reflecting high variance without tool guidance; \textit{R} reduces this variability by constraining tool selection; and \textit{H} further stabilizes performance across most dimensions. However, \textit{H} narrows variance without eliminating dimension-level gaps, suggesting that documentation improves argument construction but does not fully resolve interface complexity. At the phase level, Reconnaissance and Vulnerability Analysis are relatively stable, while Post-Exploitation and Defensive Analysis show greater variation, likely due to more heterogeneous CLI structures.

At the tool level, performance differences largely reflect CLI interface complexity. Tools with simple invocation patterns and well-defined defaults (e.g., \textit{truecrypt2john}, \textit{jsql}) are handled reliably, whereas those with irregular or uncommon options (e.g., \textit{nikto}, \textit{hashrat}) remain challenging due to precise flag--value binding and strict syntax. Additional full statistics are provided in Appendix~\ref{bestWorst}, confirming that tool-level difficulty is driven more by interface complexity than model-specific limitations.

\paragraph{Mode-wise Performance Trends.}
Figure~\ref{fig:dim}(b,c) visualizes the distributions of Total Score and Exact Correct across the three evaluation modes. Both metrics improve consistently from \textit{U} to \textit{R} to \textit{H}, with \textit{H} achieving the strongest results. This supports our earlier finding that tool documentation provides the largest gains, primarily by improving argument prediction, the main bottleneck in accurate command construction.

The two metrics exhibit different patterns of cross-model dispersion. The Total Score range narrows from 21.8 $\rightarrow$ 18.4 $\rightarrow$ 13.7 percentage points across
\textit{U}, \textit{R}, and \textit{H}.
In contrast, the Exact Correct range remains wide at 28.8, 33.8, and 28.3 percentage points, respectively. This suggests that providing tool
information brings models closer together in Total Score, while substantial differences in exact command correctness persist. 

% The two metrics show different saturation patterns. Total Score exhibits a shrinking range across modes (21$\rightarrow$15$\rightarrow$9), indicating reduced variance and suggesting that component-level performance becomes more stable with additional tool information. In contrast, Exact Correct retains a large and increasing range (21$\rightarrow$22$\rightarrow$28), indicating that perfect command reconstruction remains challenging for some models and far from saturated even in \textit{H} mode.

\paragraph{Common Failure Pattern Analysis.}
Weak query-to-command performance primarily stems from accumulated errors in command construction, particularly with optional arguments. Across models, we observe four recurring patterns: \textit{(i) long--short option misuse}, where alias forms are mixed (e.g., \verb|--|domain/\verb|-|d), producing invalid commands; \textit{(ii) option hallucination}, where positional arguments are treated as flags or nonexistent aliases inferred (e.g., `st \verb|--|client'); \textit{(iii) hallucinated long options}, where tools with only short flags lead to invented long forms;
% (e.g., `\verb|--|help'); 
and \textit{(iv) erroneous option names}, where singular--plural confusion produces incorrect variants (e.g., \verb|--|symlink vs. \verb|--|symlinks).

\begin{wraptable}{r}{0.42\linewidth}
      \centering
      \vspace{-10pt}
      \caption{Average change in Total Score, in percentage points, relative to original queries across eight model configurations.
      }
      \label{tab:query_robustness}

      \small
      \setlength{\tabcolsep}{4pt}
      \renewcommand{\arraystretch}{1.08}
      \begin{tabular}{@{}lrrr@{}}
          \toprule
          \textbf{Query style}
          & \textbf{Ures.}
          & \textbf{Res.}
          & \textbf{Hinted} \\
          \midrule
          Paraphrase & +0.61 & +0.22 & -0.09 \\
          Informal   & -0.55 & -0.48 & -0.28 \\
          Messy      & -3.20 & -1.40 & -1.07 \\
          \bottomrule
      \end{tabular}
      \vspace{-10pt}
  \end{wraptable}

% \begin{wrapfigure}{r}{0.42\linewidth}
%     \centering
%     \vspace{-10pt}
%     \includegraphics[width=\linewidth]{image/fig2_delta_vs_clean_en.pdf}
%     \caption{
%     Query Variation Robustness.
%     }
%     \label{fig:robust}
%     \vspace{-12pt}
% \end{wrapfigure}
\paragraph{Robustness to Query Variation.}
Since KaliBench queries are synthesized from tool documentation, we examine the sensitivity of model performance to changes in query wording and presentation. We evaluate eight model configurations under three query variations: \textit{Paraphrase} (uses synonyms and alternative phrasing); \textit{Informal} (adopts conversational wording); and \textit{Messy} (introduces linguistic and formatting noise, such as grammatical errors, omitted prepositions or verbs, and incorrect word choices).

Table~\ref{tab:query_robustness} shows that paraphrases have little overall effect and slightly improve performance in the \textit{Unrestricted} and \textit{Restricted} settings. Informal queries introduce a small but consistent degradation. Messy queries cause the largest drop across all settings, especially in the \textit{Unrestricted} setting. \textit{Hinted} mode shows the smallest decline under messy inputs (1.07 percentage points), suggesting that access to full tool documentation mitigates, but does not eliminate, the impact of noisy wording. The eight configurations and their model-level results for the \textit{Messy} variation are detailed in Appendix~\ref{app:robust}.

% Table~\ref{tab:query_robustness} shows that simple semantic paraphrases are largely harmless and can even slightly improve performance in the Unrestricted and Restricted settings. Informal queries introduce a small but consistent degradation. Messy queries cause the largest drop across all settings, especially in the Unrestricted setting. Hinted mode shows the smallest decline under messy inputs (1.07 percentage points), suggesting that having access to full tool documentation mitigates, but does not eliminate, the impact of noisy wording. More model-wise details can be found in Appendix~\ref{app:robust},

\section{Discussion and Conclusion}

KaliBench exposes a clear gap between LLMs’ general cybersecurity knowledge and their ability to generate exact, executable CLI commands. Our results show that tool selection and argument construction are fundamentally different challenges: restricting candidate tools largely resolves tool-selection errors, while argument-level correctness, especially for optional flags and flag–value bindings, remains the primary bottleneck. Providing tool documentation results in the largest gains, indicating that failures often stem from limited domain knowledge rather than intent understanding. These findings highlight the importance of evaluating tool use at the command-structure level in schema-free settings, where models must infer tools and syntax without explicit APIs.

Beyond evaluation, KaliBench enables effective model improvement through fine-grained, runtime-free verifiable rewards. Post-training with SFT and RLVR substantially improves compact models, allowing an 8B model to approach much larger systems. Despite these advances, limitations remain, including the focus on single-turn command generation and reliance on documentation-grounded data. 
Future work should extend to multi-step workflows, retrieval-augmented tool use, and environment-aware evaluation. 
Overall, KaliBench provides a practical benchmarking and training framework for advancing the reliable and executable use of LLM-based cybersecurity tools.

\section{Acknowledgements}
This research was funded by Khalifa University
of Science and Technology through the Faculty Start-Ups under Project ID: KU-INT-FSU-2005-8474000775. This research was also supported by KU Research Center for Advanced Intelligent Systems (AIS), Khalifa University of Science and Technology (KU-AIS).
% \NS{TODO: Finalize discussion after other sections done.}

% KaliBench isolates the translation of natural-language analyst intent into concrete CLI tool invocations, bridging the gap between knowledge-based benchmarks and fully agentic evaluations. Our results indicate that failures in argument construction, including flag semantics and CLI syntax, remain the primary barrier to correct end-to-end command generation. Performance differences are driven more by interface regularity than by cybersecurity phase, with tools exposing simple option structures handled more reliably than those with dense or irregular interfaces. These observations suggest that improving NL--to--CLI performance depends largely on better representation and interface alignment, motivating future work on documentation-aware training, constrained decoding, and system-level validation or repair.

% \section{Conclusion}

% \NS{TODO: Finalize conclusion after other sections done.}
% We introduced KaliBench, a manuscript-grounded benchmark for NL--to--CLI translation on Kali Linux, comprising 4.1K query--command pairs over 2{,}138 tools across 23 capability dimensions and five phases. Evaluations of 12 general-purpose and security-focused LLMs show that while tool selection improves with hints, end-to-end command correctness remains limited due to errors in optional-argument formulation and irregular CLI interfaces. KaliBench offers a reproducible testbed for studying cybersecurity function calling and identifying future research challenges.
%%%%%%%%%%%%%%%%%%%%%%%%%%%%

\bibliographystyle{plainnat}
\bibliography{custom.bib}

% \section*{References}

% References follow the acknowledgments in the camera-ready paper. Use unnumbered first-level heading for
% the references. Any choice of citation style is acceptable as long as you are
% consistent. It is permissible to reduce the font size to \verb+small+ (9 point)
% when listing the references.
% Note that the Reference section does not count towards the page limit.
% \medskip

% {
% \small

% [1] Alexander, J.A.\ \& Mozer, M.C.\ (1995) Template-based algorithms for
% connectionist rule extraction. In G.\ Tesauro, D.S.\ Touretzky and T.K.\ Leen
% (eds.), {\it Advances in Neural Information Processing Systems 7},
% pp.\ 609--616. Cambridge, MA: MIT Press.

% [2] Bower, J.M.\ \& Beeman, D.\ (1995) {\it The Book of GENESIS: Exploring
%   Realistic Neural Models with the GEneral NEural SImulation System.}  New York:
% TELOS/Springer--Verlag.

% [3] Hasselmo, M.E., Schnell, E.\ \& Barkai, E.\ (1995) Dynamics of learning and
% recall at excitatory recurrent synapses and cholinergic modulation in rat
% hippocampal region CA3. {\it Journal of Neuroscience} {\bf 15}(7):5249-5262.
% }

\newpage
\section*{NeurIPS Paper Checklist}

\begin{enumerate}

\item {\bf Claims}
    \item[] Question: Do the main claims made in the abstract and introduction accurately reflect the paper's contributions and scope?
    \item[] Answer: \answerYes{}% Replace by \answerYes{}, \answerNo{}, or \answerNA{}.
    \item[] Justification: Our paper's \textbf{Abstract} and \textbf{Section 1: Introduction} clearly define the article's contributions and scope.

\item {\bf Limitations}
    \item[] Question: Does the paper discuss the limitations of the work performed by the authors?
    \item[] Answer: \answerYes{} % Replace by \answerYes{}, \answerNo{}, or \answerNA{}.
    \item[] Justification: We carefully discuss the limitations of our work in \textbf{Appendix \ref{limitations}: Limitations}.

\item {\bf Theory Assumptions and Proofs}
    \item[] Question: For each theoretical result, does the paper provide the full set of assumptions and a complete (and correct) proof?
    \item[] Answer: \answerNA{} % Replace by \answerYes{}, \answerNo{}, or \answerNA{}.
    \item[] Justification: The contributions of this paper focus on experimental results rather than theoretical derivations and theoretical results.

    \item {\bf Experimental result reproducibility}
    \item[] Question: Does the paper fully disclose all the information needed to reproduce the main experimental results of the paper to the extent that it affects the main claims and/or conclusions of the paper (regardless of whether the code and data are provided or not)?
    \item[] Answer: \answerYes{} % Replace by \answerYes{}, \answerNo{}, or \answerNA{}.
    \item[] Justification: We disclose the full code and data in the link of our project page (\textbf{Abstract}). In addition, the experiments in this paper evaluate multiple open-source models and utilize the vLLM engine during the inference phase; these settings ensure the standardization and reproducibility of the experiments (\textbf{Section 4: Experiments}). More specifically, \textbf{Appendix B: CLI Command Generation} and \textbf{Appendix C: Prompt Templates} provide all the prompts needed to reproduce the generation of the benchmark data, while \textbf{Appendix E: Terminal verification Details} provides the rules for sample-wise terminal verification, and \textbf{Appendix \ref{metric_cal}: Details about Metrics Calculation} details the metric calculations required to accurately reproduce the result. 

\item {\bf Open access to data and code}
    \item[] Question: Does the paper provide open access to the data and code, with sufficient instructions to faithfully reproduce the main experimental results, as described in supplemental material?
    \item[] Answer: \answerYes{} % Replace by \answerYes{}, \answerNo{}, or \answerNA{}.
    \item[] Justification: We provide full access to the data (via Hugging Face) and the complete implementation code (via GitHub). Link contains these resources can be found at the beginning of the paper, along with sufficient instructions to reproduce the main experimental results.

\item {\bf Experimental Setting/Details}
    \item[] Question: Does the paper specify all the training and test details (e.g., data splits, hyperparameters, how they were chosen, type of optimizer) necessary to understand the results?
    \item[] Answer: \answerYes{} % Replace by \answerYes{}, \answerNo{}, or \answerNA{}.
    \item[] Justification: 
    The dataset splits, evaluation protocol, and training methods are described in \textbf{Section 3}. Model configurations and detailed training hyperparameters are provided in \textbf{Appendix G}.
    %We specify all the training \justificationTODO{} and test details (\textbf{Section 4 Experiments}).

\item {\bf Experiment Statistical Significance}
    \item[] Question: Does the paper report error bars suitably and correctly defined or other appropriate information about the statistical significance of the experiments?
    \item[] Answer: \answerYes{} % Replace by \answerYes{}, \answerNo{}, or \answerNA{}.
    \item[] Justification: The paper provides information about variability through distributional statistics. In particular, \textbf{Figure 5} reports box plots (Q1–Q3 with min/max whiskers) across all models, security dimensions, and mode settings.
    %, which illustrate the spread and variability of the results. While formal confidence intervals or hypothesis tests are not included, these distributional summaries provide useful insight into the stability and variation of the experimental outcomes.

\item {\bf Experiments Compute Resources}
    \item[] Question: For each experiment, does the paper provide sufficient information on the computer resources (type of compute workers, memory, time of execution) needed to reproduce the experiments?
    \item[] Answer: \answerYes{} % Replace by \answerYes{}, \answerNo{}, or \answerNA{}.
    \item[] Justification: We have provided the GPU model, memory, and the amount of compute required for the experiment (\textbf{Section 4: Experiments, Appendix \ref{appendix:training-protocol}}). The full research project required the same compute as reported in the paper.
    
\item {\bf Code Of Ethics}
    \item[] Question: Does the research conducted in the paper conform, in every respect, with the NeurIPS Code of Ethics \url{https://neurips.cc/public/EthicsGuidelines}?
    \item[] Answer: \answerYes{} % Replace by \answerYes{}, \answerNo{}, or \answerNA{}.
    \item[] Justification: We carefully checked the NeurIPS Code of Ethics and discussed in \textbf{Appendix A: Ethical Statement} that this paper does not raise any ethical issues.

\item {\bf Broader Impacts}
    \item[] Question: Does the paper discuss both potential positive societal impacts and negative societal impacts of the work performed?
    \item[] Answer: \answerYes{} % Replace by \answerYes{}, \answerNo{}, or \answerNA{}.
    \item[] Justification: We discuss the potential societal impacts in \textbf{Appendix A: Ethical Statement}.
    
\item {\bf Safeguards}
    \item[] Question: Does the paper describe safeguards that have been put in place for responsible release of data or models that have a high risk for misuse (e.g., pre-trained language models, image generators, or scraped datasets)?
    \item[] Answer: \answerYes{} % Replace by \answerYes{}, \answerNo{}, or \answerNA{}.
    \item[] Justification: The paper discusses potential dual-use risks of the cybersecurity benchmark and describes several release-related precautions in \textbf{Appendix A: Ethical Statement}.

\item {\bf Licenses for existing assets}
    \item[] Question: Are the creators or original owners of assets (e.g., code, data, models), used in the paper, properly credited and are the license and terms of use explicitly mentioned and properly respected?
    \item[] Answer: \answerYes{} % Replace by \answerYes{}, \answerNo{}, or \answerNA{}.
    \item[] Justification: \textbf{Appendix A: Ethical Statement} documents the existing assets used in this work, credits their original sources, and states the applicable licenses and terms of use; we listed the evaluated model details in \textbf{Appendix \ref{appendix:evaluated-model}}.

\item {\bf New assets}
    \item[] Question: Are new assets introduced in the paper well documented and is the documentation provided alongside the assets?
    \item[] Answer: \answerYes{} % Replace by \answerYes{}, \answerNo{}, or \answerNA{}.
    \item[] Justification: The paper introduces KaliBench as a new benchmark asset, and \textbf{Appendix A: Ethical Statement} documents its construction, intended use, licensing, and release considerations. Besides, Section 3 describes the benchmark's construction, data structure, verification, splits, and evaluation protocol.

\item {\bf Crowdsourcing and research with human subjects}
    \item[] Question: For crowdsourcing experiments and research with human subjects, does the paper include the full text of instructions given to participants and screenshots, if applicable, as well as details about compensation (if any)?     
    \item[] Answer: \answerNA{} % Replace by \answerYes{}, \answerNo{}, or \answerNA{}.
    \item[] Justification: \textbf{Appendix A: Ethical Statement} states that no data were collected from human subjects and no external annotators were involved; the paper does not include crowdsourcing or human-subject research.

\item {\bf Institutional review board (IRB) approvals or equivalent for research with human subjects}
    \item[] Question: Does the paper describe potential risks incurred by study participants, whether such risks were disclosed to the subjects, and whether Institutional Review Board (IRB) approvals (or an equivalent approval/review based on the requirements of your country or institution) were obtained?
    
    \item[] Answer: \answerNA{} % Replace by \answerYes{}, \answerNo{}, or \answerNA{}.
    \item[] Justification: \textbf{Appendix A: Ethical Statement} states that no data were collected from human subjects and no external annotators were involved; therefore, this work does not involve human-subject research or crowdsourcing.

\item {\bf Declaration of LLM usage}
    \item[] Question: Does the paper describe the usage of LLMs if it is an important, original, or non-standard component of the core methods in this research? Note that if the LLM is used only for writing, editing, or formatting purposes and does \emph{not} impact the core methodology, scientific rigor, or originality of the research, declaration is not required.
    %this research? 
    \item[] Answer: \answerYes{} % Replace by \answerYes{}, \answerNo{}, or \answerNA{}.
    \item[] Justification: The paper clearly describes the use of LLMs as a core component, including their roles in data generation, verification (LLM-as-verifier), and model training (SFT and RL with verifiable rewards), as detailed in \textbf{Section 3}. %\justificationTODO{}

\end{enumerate}
%%%%%%%%%%%%%%%%%%%%%%%%%%%%%%%%%%%%%%%%%%%%%%%%%%%%%%%%%%%%

\appendix

\section{Ethical Statement}
\noindent\textbf{Existing Assets.}
All data used in this work are derived from publicly available Kali Linux tool manuals and documentation. This source repository is part of the Kali Linux documentation project and is licensed under the GNU General Public License v3.0. Model inference was performed either offline in a controlled environment or via APIs. 
Open-weight models and proprietary API services are subject to their respective model licenses and provider terms. Except for our KaliBench post-trained variants, we did not
perform additional fine-tuning on KaliBench.

% Except for our KaliBench post-trained variants, all pretrained models evaluated in this work were used in accordance with their respective open-source licenses and without additional fine-tuning on KaliBench.

\noindent\textbf{New Assets.}
KaliBench is intended for research purposes only and is released under the CC BY-NC 4.0 license. KaliBench is released together with documentation describing the structure of the benchmark, dataset fields, data construction and verification pipeline, train--test splits, evaluation protocol, licensing terms, intended use, limitations, and responsible-use guidelines. Since the dataset is derived from publicly available Kali Linux documentation and contains no personally identifiable information or human-subject data, no individual consent is required. We also provide release notes and usage instructions to support reproducible and responsible use of the benchmark.

\paragraph{Artifact availability.}
The public repository provides code and instructions for data
generation, filtering, verification, evaluation, and
post-training, including SFT and GRPO.\footnote{
\url{https://github.com/RISys-Lab/KaliBench}}
Croissant metadata was also provided with the OpenReview
submission to support machine-readable dataset documentation.

\noindent\textbf{Human Subjects.}
The KaliBench dataset contains no personally identifiable information. No data were collected from human subjects, and no external annotators were involved in data creation or evaluation. All human-in-the-loop verification is conducted by the authors solely for internal data quality control; no external annotators, crowd workers, or human subjects were involved.

% \noindent\textbf{Annotator.}
%  The dataset is constructed from publicly available tool documentation (Kali-Linux) and does not introduce new vulnerabilities or implement end-to-end attack automation. As with many cybersecurity benchmarks, potential dual-use considerations may arise, since the tools included can be applied in both defensive and offensive contexts. By highlighting systematic failure patterns and limitations of current models, our work aims to support the safe and responsible use of LLMs in cybersecurity practice.

\noindent\textbf{Social Impact.}
The main positive societal impact of this work lies in improving the reliability evaluation of large language models in cybersecurity settings. This work helps reduce the risk that security-oriented LLMs generate incorrect commands or mislead users before real-world deployment, and can support the development of more reliable defensive tools for legitimate scenarios such as security education, vulnerability assessment, incident response, and digital forensics. This work does not introduce new attack techniques, vulnerability exploitation methods, or end-to-end attack automation workflows. The data are primarily derived from publicly available Kali Linux tool documentation. KaliBench focuses only on single-command-level translation and evaluation, and does not provide multi-step attack chains, target-environment-adaptive attack procedures, or end-to-end automated penetration capabilities. Therefore, we believe that KaliBench does not directly introduce new security threats.

\noindent\textbf{LLM Usage.}
% Parts of the manuscript were refined using large language models (e.g., ChatGPT) for language editing only; all experimental design, analysis, and conclusions were produced and verified by the authors.
LLMs were used for query--command generation,
documentation-grounded verification, and preliminary assessment of suspicious examples, with final manual-review decisions made by the authors. LLMs were also used for language editing of the manuscript. All experimental design, analysis, and conclusions were produced and verified by the authors.

\section{CLI Command Generation}\label{data generation}
% Most existing function-calling benchmarks operate over structured APIs, where argument order, spacing, and quoting rarely break execution. By contrast, cybersecurity toolchains expose \emph{rigid, syntax-sensitive} command-line interfaces: Kali Linux tools require precise binary names, alias conventions, flag--value spacing, quoting behavior, and the exact ordering of positional arguments. Even minor deviations such as an extra space, a misplaced flag, an incorrect alias can invalidate a command.

KaliBench focuses on deterministic \textit{natural-language-to-CLI} translation at scale within Kali Linux ($\sim$2,372 tools). It defines reviewer-checkable canonicalization rules that specify:

\begin{enumerate}[label=\roman*., leftmargin=*]
    \item \textbf{Ordering}: the command begins with the tool name, followed by all long flags, then all short flags (lexicographically), then flags with values, and finally positional arguments in document order.
    
    \item \textbf{Flag syntax}: flags and their values are space-separated (e.g., `\verb|--|opt \texttt{value}', `\texttt{-s value}'). Equals-style bindings (e.g., `\verb|--|opt=value') are allowed. Boolean flags take no value.
    
    \item \textbf{Quoting and aliasing}: quoting is applied only when required (e.g., values containing whitespace or shell meta-characters). Aliases (e.g., `\verb|--|help' vs.\ `\texttt{-h}') are normalized, but the model must output a single canonical form---preferably the long-flag variant.
\end{enumerate}

To ensure unambiguous extraction, all CLI predictions are enclosed within a standardized \texttt{<output>}~\texttt{</output>} tag. This enables robust isolation and symbolic comparison of the core command.

In all, these constraints enable scalable and accurate evaluation by supporting diverse arguments and alias combinations in the data generation process, ensuring reproducibility through exact parse-tree matching.\\

\section{Prompt Templates}\label{prompt}
In this section, we list the three prompt templates used in our study, corresponding to
the query-command generation, LLM-assisted data verification, and model evaluation.

\subsection{Prompt for Generating the Query-Command Pairs}
This prompt defines a controlled data generation setting for constructing a cybersecurity command benchmark from official Kali/Linux tool documentation. The model is instructed to synthesize natural-language queries that uniquely determine a single, canonical shell command, together with structured argument annotations. Strict constraints enforce faithfulness to documented tool capabilities, deterministic command formatting, and explicit parameter specification, thereby reducing ambiguity and unsupported behaviors. The resulting dataset enables reliable evaluation of command generation, argument parsing, and hallucination under reproducible conditions.

% === Prompt Instruction Box ===
\begin{tcolorbox}[promptbox, title=\textbf{System Instruction}]
``You are generating a cybersecurity tool benchmark dataset. 
Your task is to convert seed documentation of Kali/Linux tools into a structured JSON dataset... 
\end{tcolorbox}

% === Goal Definition Box ===
\begin{tcolorbox}[promptbox, title=\textbf{Goal}]
The goal is to generate a JSON array where each entry consists of: 1.An unambiguous natural language query 2.The exact ground-truth shell command 3.The canonical tool name 4.Parsed optional arguments (flag--value mapping) 5.Ordered positional arguments.
\end{tcolorbox}

% === Hard Constraints Box (Condensed) ===
\begin{tcolorbox}[promptbox, title=\textbf{Hard Constraints}]
\begin{enumerate}
    \item Use only tools, flags, and behaviors explicitly documented in the seed.
    \item Commands must follow deterministic canonicalization
    (tool $\rightarrow$ long flags $\rightarrow$ short flags $\rightarrow$ value flags $\rightarrow$ positional arguments).
    \item No invented flags, pipelines, subshells, or implicit defaults.
    \item Each query must admit exactly 1 valid command, with aliases documented.
\end{enumerate}
\end{tcolorbox}

% === Unambiguity Enforcement Box (Condensed) ===
\begin{tcolorbox}[promptbox, title=\textbf{Unambiguity Enforcement}]
To ensure a single correct answer:
\begin{itemize}
    \item All required parameters must be explicit; vague phrasing (e.g., ``quick'', ``default'') is forbidden.
    % \item If multiple tools could apply, the tool must be named.
    \item A uniqueness self-check is required (no alternative command fits the query).
\end{itemize}
\end{tcolorbox}

% === Output Format Box ===
\begin{tcolorbox}[promptbox, title=\textbf{Output Format}]
The output must be a JSON array, where each item strictly follows:
\begin{verbatim}
{
  "query": "string",
  "ground_truth_command": "string",
  "tool_name": "string",
  "optional_args": { "flag|alias": 
           "value or null" },
  "positional_args": ["arg1", "arg2"]
}
\end{verbatim}
No additional text is permitted.
\end{tcolorbox}

\subsection{Prompt for LLM-assisted Data Verification} 
\label{appendix:subsec-llm-verification}

The following prompt is designed for dataset cleaning and validation by instructing the model to act as a strict CLI command verifier. Given a <natural language query>, a <corresponding ground-truth command>, and <the official usage documentation of a specified tool>, the model evaluates whether the command is valid or hallucinated. The verification focuses on flag existence, argument correctness, and functional alignment with the documented capabilities of the tool. The output consists of a binary verdict accompanied by a concise justification, enabling reliable filtering of erroneous or unsupported command annotations.

% === Prompt Instruction Box ===
\begin{tcolorbox}[promptbox, title=\textbf{System Instruction}]
``You are a meticulous CLI command verifier. Verify whether the query--ground truth pairs 
(\texttt{query}, \texttt{ground\_truth\_command}) are accurate or hallucinated based on the provided <Tool Documentation>.''
\end{tcolorbox}

% === Verification Criteria Box ===
\begin{tcolorbox}[promptbox, title=\textbf{Verification Criteria}]
The model must carefully check:
\begin{enumerate}
    \item Whether every flag and argument in the ground-truth command appears in the tool usage documentation and is used correctly.
    \item Whether the command can reasonably implement the requested query using only the capabilities described in the usage documentation.
    \item If any flag does not exist or the requested behavior is unsupported, the command should be classified as hallucinated.
\end{enumerate}
\end{tcolorbox}

% === Output Format Box ===
\begin{tcolorbox}[promptbox, title=\textbf{Output Format}]
The model must reason step-by-step and produce the following structured output:
\begin{verbatim}
### Reasoning
<Step-by-step reasoning>

### Output JSON
{
"verdict": "ACCURATE"/"HALLUCINATED",
"reasons": "<short explanation>"
}
\end{verbatim}
\end{tcolorbox}

\subsection{Prompt for Model Evaluation} \label{evaluation_prompt}
The following prompt tells the model how to analyze a query and generate corresponding CLI commands in \textbf{\textit{Unrestricted}} mode for \textbf{evaluation}. The model is instructed to act as a cybersecurity function-calling agent with direct access to a Kali/Linux execution environment. Given a user query, the model is required to output a single, executable shell command that precisely accomplishes the requested task. The prompt enforces strict constraints on tool selection, argument completeness, and option–value correctness, while explicitly prohibiting explanatory text or non-command output. This design enables controlled evaluation of the model’s accuracy and completeness in generating real-world security-related commands. System prompts for \textbf{\textit{Restricted}} and \textit{\textbf{Hinted}} modes are similar as that of unrestricted. The only difference is that these two modes embed a subset of <{\textit{``a group of tool names"}}> and <{\textit{``a group of tool names with their corresponding usage manuscripts"}}> in the Task Definition Part.
% === Prompt Instruction Box ===
\begin{tcolorbox}[colback=myBG, colframe=black,
                  arc=2mm, boxrule=0.5pt,
                  left=2mm,right=2mm,top=1mm,bottom=1mm,
                  breakable, title=\textbf{System Instruction}]
``You are a cybersecurity function-calling AI model.
You have access to the
\texttt{run\_terminal}. Execute a shell command in a Kali/Linux terminal and return stdout, stderr, and exit code.''
\end{tcolorbox}

% === Task Definition Box ===
\begin{tcolorbox}[colback=myBG, colframe=black,
                  arc=2mm, boxrule=0.5pt,
                  left=2mm,right=2mm,top=1mm,bottom=1mm,
                  breakable, title=\textbf{Task Definition}]
``Given a <USER QUERY>, generate the single most accurate shell command using any appropriate Kali/Linux tool(s) to solve the query.''
\end{tcolorbox}

% === Command Generation Constraints Box ===
\begin{tcolorbox}[colback=myBG, colframe=black,
                  arc=2mm, boxrule=0.5pt,
                  left=2mm,right=2mm,top=1mm,bottom=1mm,
                  breakable, title=\textbf{Command Generation Constraints}]
1. Use the correct command-line tool(s) appropriate for the task. \\
2. Include all required optional arguments (flags beginning with \verb|-| or \verb|--|). \\
3. Correctly pair option keys and values (e.g., \verb|--port 80|, \verb|-A INPUT|). \\
4. Preserve all positional arguments (e.g., IP addresses, filenames, interfaces). \\
5. Do \textbf{not} invent flags or options that do not exist in real Kali/Linux tools. \\
6. Do \textbf{not} include explanations, commentary, or extraneous text. \\
7. Missing flags, incorrect option-value pairs, or omitted positional arguments are penalized.
\end{tcolorbox}

% === Output Format Box ===
\begin{tcolorbox}[colback=myBG, colframe=black,
                  arc=2mm, boxrule=0.5pt,
                  left=2mm,right=2mm,top=1mm,bottom=1mm,
                  breakable, title=\textbf{Output Format}]
\texttt{<output>} \\
\texttt{[command]} \\
\texttt{</output>}
\end{tcolorbox}

\section{KaliBench Samples in Detail} \label{text}

\begin{figure}[t]
    \centering
    \includegraphics[width=\linewidth]{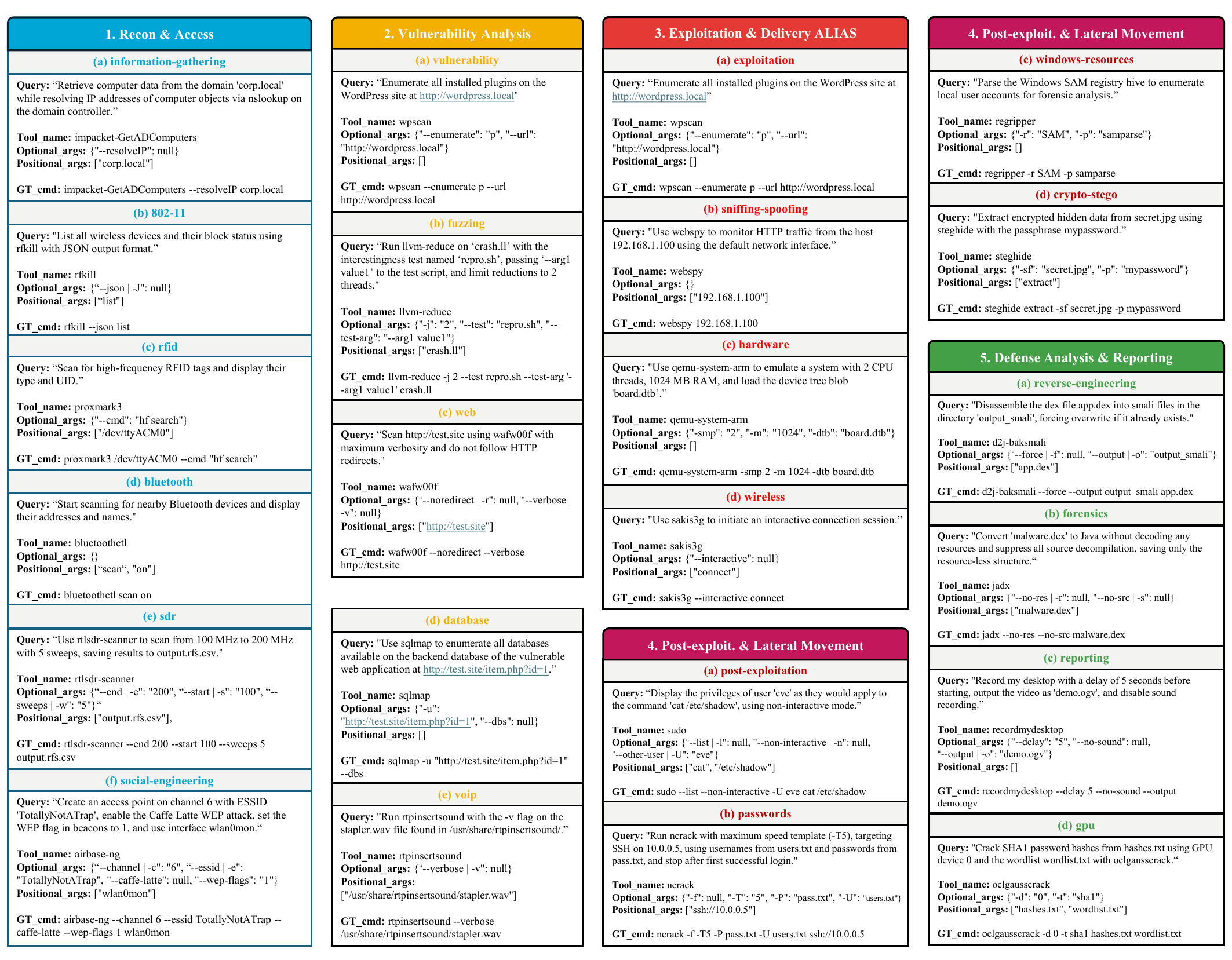}
    \caption{
    \textbf{Overview of KaliBench samples.}
    Representative examples spanning all five security phases and 23 evaluation dimensions are shown. Zoom in for finer details.
    }
    \label{fig:kalibench_overview}
\end{figure}

This section provides a detailed overview of KaliBench. 
As illustrated in Figure~\ref{fig:kalibench_overview}, the benchmark is structured around five core security phases, covering 23 fine-grained evaluation dimensions. 
Each instance consists of a natural language query, its corresponding tool invocation (including optional and positional arguments), and a ground-truth command. 
This design enables systematic evaluation of a model’s ability to interpret, decompose, and execute security-oriented tasks. 
Overall, KaliBench captures a diverse range of real-world security workflows.

\section{Terminal verification Details} \label{terminal}
We perform sample-wise terminal verification by executing each \texttt{ground\_truth\_command} in a controlled Docker sandbox (with Kali Linux environment). For each command, we create an isolated working directory to avoid side effects, and run it via \texttt{subprocess} with a fixed timeout and resource limits. During execution, we capture \texttt{stdout}, \texttt{stderr}, \texttt{exit\_code}, and runtime signals.

After execution, we assign labels based on simple rules:
\begin{itemize}
\item \textbf{PASS\_EXECUTED}: exit code = 0
\item \textbf{TIMEOUT}: command exceeds time limit
\item \textbf{REJECT\_HALLUCINATED\_TOOL}: tool not found (e.g., exit code 127 or error patterns)
\item \textbf{REJECT\_HALLUCINATED\_PARAMETER}: invalid or unsupported options detected from error messages
\item \textbf{PASS\_RUNTIME\_ERROR}: non-zero exit code without clear hallucination signals
\end{itemize}

We use regex-based pattern matching on \texttt{stderr/stdout} to detect tool and parameter errors, ensuring that obvious hallucinations can be filtered out even if the command executes.

\textbf{A brief Pseudo-code can further show our execution process:}
\begin{verbatim}
# Input: JSONL dataset of query–command pairs
# Output: execution logs with classification labels
for each row in dataset:
    # 1. Prepare command
    cmd = row["ground_truth_command"]

    # 2. Create isolated sandbox
    workdir = create_sandbox(cmd)

    # 3. Execute command with timeout
    result = run(cmd, timeout=20s, in workdir)
    stdout, stderr = result.stdout, result.stderr
    exit_code = result.exit_code

    # 4. Classify execution outcome
    if result.timed_out:
        label = TIMEOUT

    elif exit_code == 127 or match(tool_error_pattern, stderr, stdout):
        label = REJECT_HALLUCINATED_TOOL

    elif match(parameter_error_pattern, stderr, stdout):
        label = REJECT_HALLUCINATED_PARAMETER

    elif exit_code == 0:
        label = PASS_EXECUTED

    else:
        label = PASS_RUNTIME_ERROR

    # 5. Save execution log
    save {
        command, stdout, stderr,
        exit_code, label, workdir
    }
\end{verbatim}

This process filters out commands with hallucinated tools or invalid parameters, while retaining executable commands and partially valid ones that are \textbf{syntactically correct but may fail at runtime due to placeholder inputs} (e.g., using non-existent targets such as \texttt{curl https://example.com}).

\section{Representative Invalid Cases Identified During Manual Review}
\label{manual case}

To further substantiate the error types summarized in the main text, we provide representative invalid cases identified during manual review. 

\paragraph{(a) Malformed CLI options.}
For the query ``Launch \texttt{debug\_me} via \texttt{qemu-aarch64\_be} and wait for a GDB connection on port 1234,'' the correct command is ``\texttt{qemu-aarch64\_be -g 1234 debug\_me}'', whereas the annotated ground truth uses the malformed option \texttt{\dashdash g} instead of \texttt{-g}.

\paragraph{(b) Omission of query-required arguments.}
For the query ``Generate shell completion script for \texttt{kustomize} so it can be used with \texttt{bash},'' the correct command is ``\texttt{kustomize completion bash}'', but the annotated ground truth uses only ``\texttt{kustomize completion}'' and omits the shell argument, which is specifically asked by the query.

\paragraph{(c) Semantic mismatches in argument use.}
We observe three common subtypes. First, the annotated command may use a flag with the wrong function: for the query ``Run Git with the working directory set to \texttt{/home/user/project},'' the correct command is ``\texttt{git -C /home/user/project status}'', while the ground truth uses \texttt{\-\-work-tree}, which changes the repository work tree rather than the working directory. Second, the annotated command may reverse operand order: for the query ``converting \texttt{mac.txt} to \texttt{unix3.txt}'', the correct form is \texttt{dos2unix -n mac.txt unix3.txt}, while the ground truth swaps input and output. Third, the annotated command may use the wrong argument type: for the query "Execute svnsync's 'sync' subcommand on the local mirror located at '/srv/svn/replica'", the correct command is "svnsync sync file:///srv/svn/replica", while the ground truth supplies the plain path ``\texttt{/srv/svn/replica}'' where a URL is required.

% \paragraph{(d) Other low-frequency but representative errors.}
% Finally, manual review reveals several lower-frequency but still representative annotation problems. One example is query--tool inconsistency: for the query ``Compile \texttt{program.c} with \texttt{afl-clang-fast++},'' the explanation in \texttt{GT\_false} indicates that the C source file is more appropriately handled by the C driver \texttt{afl-clang-fast}, suggesting that the query itself conflicts with the intended tool variant. Another example is omission of required execution context: for the query requiring \texttt{rsmangler} to consume input via \texttt{cat words.txt |}, the annotated ground truth omits the required stdin feed entirely. We also observe rare outright false negatives, such as a case where the \texttt{GT\_false} explanation explicitly states that the annotated ground-truth command correctly satisfies the query, indicating a direct verdicting or annotation mistake rather than a genuine command mismatch.

% We also find lower-frequency cases such as query--tool inconsistency (e.g., a query naming \texttt{afl-clang-fast++} for a \texttt{.c} source file), missing required execution context (e.g., omitting the \texttt{cat words.txt |} stdin feed required by the query), and rare outright false negatives where the \texttt{GT\_false} explanation itself states that the annotated ground-truth command is correct.

\section{Experiment Details}

\subsection{Evaluated Model Details.}
\label{appendix:evaluated-model}

We evaluated open-source models with diverse sizes and architectures from 8B to 753B parameters. Models were either run locally using weights obtained from their official Hugging Face repositories or accessed via APIs. Table~\ref{tab:evaluated-models} summarizes the evaluated models, their parameter counts, inference modes, and Hugging Face repositories.

We used a fixed temperature of 0.2 and a maximum context length of 16,384 tokens, reserving 8,192 tokens for the input and 8,192 tokens for the output. When the combined prompt and tool documentation exceeded the reserved input budget, the tool documentation was truncated to fit within the input limit. Models with fewer than 72B parameters were evaluated on our $4 \times$ A100 GPU server using the vLLM inference engine with BF16 precision. Models with 72B or more parameters
were evaluated through the OpenRouter API.

\begin{table}[t]
\centering
\small
\setlength{\tabcolsep}{4pt}
\renewcommand{\arraystretch}{1.08}
\begin{tabularx}{\linewidth}{lclX}
\toprule
\textbf{Model Name} & \textbf{Size} & \textbf{Inference} & \textbf{Hugging Face Repository} \\
\midrule
Llama3.1-Instruct & 8B & local vLLM & \href{https://huggingface.co/meta-llama/Llama-3.1-8B-Instruct}{meta-llama/Llama-3.1-8B-Instruct} \\
Foundation-Sec-Instruct & 8B & local vLLM & \href{https://huggingface.co/fdtn-ai/Foundation-Sec-8B}{fdtn-ai/Foundation-Sec-8B} \\
Foundation-Sec-Reasoning & 8B & local vLLM & \href{https://huggingface.co/fdtn-ai/Foundation-Sec-8B-Reasoning}{fdtn-ai/Foundation-Sec-8B-Reasoning} \\
Llama-Primus-Base & 8B & local vLLM & \href{https://huggingface.co/trend-cybertron/Llama-Primus-Base}{trend-cybertron/Llama-Primus-Base} \\
Llama-Primus-Merged & 8B & local vLLM & \href{https://huggingface.co/trendmicro-ailab/Llama-Primus-Merged}{trendmicro-ailab/Llama-Primus-Merged} \\
RedSage-Qwen3-8B-Ins & 8B & local vLLM & \href{https://huggingface.co/RISys-Lab/RedSage-Qwen3-8B-Ins}{RISys-Lab/RedSage-Qwen3-8B-Ins} \\
RedSage-Qwen3-8B-DPO & 8B & local vLLM & \href{https://huggingface.co/RISys-Lab/RedSage-Qwen3-8B-DPO}{RISys-Lab/RedSage-Qwen3-8B-DPO} \\
RedSage-K (SFT) & 8B & local vLLM &
\href{https://huggingface.co/RISys-Lab/RedSage-K-SFT}
{RISys-Lab/RedSage-K-SFT} \\

RedSage-K (GRPO) & 8B & local vLLM &
\href{https://huggingface.co/RISys-Lab/RedSage-K-GRPO}
{RISys-Lab/RedSage-K-GRPO} \\

RedSage-K (SFT+GRPO) & 8B & local vLLM &
\href{https://huggingface.co/RISys-Lab/RedSage-K-SFT-GRPO}
{RISys-Lab/RedSage-K-SFT-GRPO} \\

%RedSage-K (SFT) & 8B & local vLLM & To be released \\
%RedSage-K (GRPO) & 8B & local vLLM & To be released \\
%RedSage-K (SFT+GRPO) & 8B & local vLLM & To be released \\

Qwen-3-8B & 8B & local vLLM & \href{https://huggingface.co/Qwen/Qwen3-8B}{Qwen/Qwen3-8B} \\
GPT-OSS-20B & 21B & local vLLM & \href{https://huggingface.co/openai/gpt-oss-20b}{openai/gpt-oss-20b} \\
Mistral-3.2-24B & 24B & local vLLM & \href{https://huggingface.co/mistralai/Mistral-Small-3.2-24B-Instruct-2506}{mistralai/Mistral-Small-3.2-24B-Instruct-2506} \\
Gemma-3-27B & 27B & local vLLM & \href{https://huggingface.co/google/gemma-3-27b-it}{google/gemma-3-27b-it} \\
Qwen-3-32B & 32B & local vLLM & \href{https://huggingface.co/Qwen/Qwen3-32B}{Qwen/Qwen3-32B} \\
LLaMA-3.3-Instruct & 70B & local vLLM & \href{https://huggingface.co/meta-llama/Llama-3.3-70B-Instruct}{meta-llama/Llama-3.3-70B-Instruct} \\
Llama-Primus-Nemotron-70B-Instruct & 71B & local vLLM & \href{https://huggingface.co/trend-cybertron/Llama-Primus-Nemotron-70B-Instruct}{trend-cybertron/Llama-Primus-Nemotron-70B-Instruct} \\
Qwen-2.5-Instruct & 72B & OpenRouter API & \href{https://huggingface.co/Qwen/Qwen2.5-72B-Instruct}{Qwen/Qwen2.5-72B-Instruct} \\
Qwen3-Coder-Next & 80B & OpenRouter API & \href{https://huggingface.co/Qwen/Qwen3-Coder-Next}{Qwen/Qwen3-Coder-Next} \\
GPT-OSS-120B & 120B & OpenRouter API & \href{https://huggingface.co/openai/gpt-oss-120b}{openai/gpt-oss-120b} \\
DeepSeek-V3.2 & 685B & OpenRouter API & \href{https://huggingface.co/deepseek-ai/DeepSeek-V3.2}{deepseek-ai/DeepSeek-V3.2} \\
GLM-5.2 & 753B & OpenRouter API & \href{https://openrouter.ai/z-ai/glm-5.2}{zai-org/GLM-5.2} \\
\bottomrule
\end{tabularx}
\caption{
Evaluated open-weight model configurations,
parameter counts, inference platforms, and model sources. The inference platform used for each configuration is specified in the table.}

% Evaluated open-source models with parameter counts, inference modes, and Hugging Face repositories. Models larger than 72B parameters were evaluated through the OpenRouter API; all other models were evaluated locally with vLLM.

\label{tab:evaluated-models}
\end{table}

\subsection{Training Setup.}\label{appendix:training-protocol}
We fine-tuned \texttt{RedSage-Qwen3-8B-Ins} using parameter-efficient LoRA adapters on a single NVIDIA H200 GPU. For both SFT and GRPO training, we used LoRA rank 64, LoRA alpha 128, and zero LoRA dropout, applying adapters to the attention and MLP projection layers: \texttt{q\_proj}, \texttt{k\_proj}, \texttt{v\_proj}, \texttt{o\_proj}, \texttt{gate\_proj}, \texttt{up\_proj}, and \texttt{down\_proj}. Training used bfloat16 precision, gradient checkpointing, and the 8-bit AdamW optimizer with learning rate $5\times10^{-6}$, weight decay $10^{-3}$, Adam betas $(0.9, 0.999)$, gradient clipping at 1.0, and a linear learning-rate schedule with 10\% warmup. We trained for 2 epochs. The software stack included Transformers 4.57.6, Unsloth 2026.4.6, PEFT 0.19.1, and Python 3.12.13.

For the SFT model, we trained from the \texttt{RedSage-Qwen3-8B-Ins} checkpoint with a per-device batch size of 4 and gradient accumulation of 8, giving an effective batch size of 32. The maximum sequence length was 16,384 tokens, with packing disabled and padding-free training enabled. The SFT objective used the standard negative log-likelihood loss. SFT training required $\sim$32 minutes.

For the GRPO-trained model, we used the same optimizer, LoRA, precision, and batch settings as SFT. GRPO used 8 generations per prompt, a generation batch size of 32, and group-based reward scaling. The maximum prompt length was 5,549 tokens, and the maximum completion length was 6,739 tokens. We used the BNPO-style GRPO loss with clipping parameter $\epsilon=0.2$ and KL coefficient $\beta=0.001$. Generation during GRPO was accelerated with colocated vLLM, using tensor parallel size 1 and GPU memory utilization 0.3. GRPO training required $\sim$28 hours when initialized directly from \texttt{RedSage-Qwen3-8B-Ins}, and $\sim$17 hours when initialized from our SFT checkpoint.

% \NS{TODO: Complete this.}

% \SC{We need training settings as required by Checklist-6, add it here}

% \SC{We need GPU memory cost, and the amount of compute (GPU hours) as required by Checklist-8}
\begin{table}[t]
\centering
\caption{\textbf{Performance deltas across execution modes.}
We report differences between three execution modes: Unrestricted (U), Restricted (R), and Hinted (H). $\Delta_1$ denotes (R$-$U), $\Delta_2$ denotes (H$-$R), and $\Delta_3$ denotes (H$-$U), all measured in percentage points (pp). $^{\dagger}$ indicates thinking-enabled models, and $^{\star}$ denotes cybersecurity-tuned models. \textbf{Bold}/\underline{underline} mark the largest and second-largest deltas within each size group. Model order follows Table~\ref{tab:combined_perf}.}
\label{tab:deltas}
\resizebox{0.98\linewidth}{!}{%
\begin{tabular}{l | c|
    r r r
    r r r
    r r r
    r r r|
    r r r
}
\toprule
\rowcolor{blue!10}
\multirow{2}{*}{\textbf{Model}} & \multirow{2}{*}{\textbf{Size}} &
\multicolumn{3}{c}{\textbf{Tool score $\Delta$ (pp)}} &
\multicolumn{3}{c}{\textbf{Optional F1 $\Delta$ (pp)}} &
\multicolumn{3}{c}{\textbf{Positional F1 $\Delta$ (pp)}} &
\multicolumn{3}{c}{\textbf{Total score $\Delta$ (pp)}} &
\multicolumn{3}{c}{\textbf{Exact $\Delta$ (pp)}} \\
\cmidrule(lr){3-5} \cmidrule(lr){6-8} \cmidrule(lr){9-11} \cmidrule(lr){12-14} \cmidrule(lr){15-17}
 & & $\Delta_1$ & $\Delta_2$ & $\Delta_3$
   & $\Delta_1$ & $\Delta_2$ & $\Delta_3$
   & $\Delta_1$ & $\Delta_2$ & $\Delta_3$
   & $\Delta_1$ & $\Delta_2$ & $\Delta_3$
   & $\Delta_1$ & $\Delta_2$ & $\Delta_3$ \\
\midrule
 
\multicolumn{17}{>{\columncolor{gray!20}}c}{\textbf{Compact Models ($\leq$ 8B)}} \\
\midrule
Foundation-Sec-Ins$^{\star}$ & 8
& 26.00 & -5.70 & 20.30
& 3.90 & 38.20 & 42.10
& 7.00 & 21.30 & 28.30
& 12.30 & 17.90 & 30.20
& 4.40 & 37.10 & 41.50 \\

Llama3.1-Ins & 8
& \textbf{32.00} & 1.70 & \textbf{33.70}
& \underline{4.80} & \underline{45.50} & \underline{50.30}
& \underline{8.40} & 23.90 & 32.30
& \textbf{15.10} & 23.70 & \textbf{38.80}
& 5.70 & 44.50 & 50.20 \\

Llama-Primus-Base$^{\star}$ & 8
& 23.40 & \textbf{3.40} & 26.80
& 2.60 & 41.60 & 44.20
& 5.60 & 22.10 & 27.70
& 10.50 & 22.40 & 32.90
& 3.50 & 41.00 & 44.50 \\

Foundation-Sec-Rsn$^{\star}$$^{\dagger}$ & 8
& 25.80 & \underline{3.10} & 28.90
& 4.30 & \textbf{46.40} & \textbf{50.70}
& 4.90 & \textbf{28.30} & \underline{33.20}
& 11.70 & \textbf{25.90} & \underline{37.60}
& 5.30 & 48.00 & 53.30 \\

Llama-Primus-Merged$^{\star}$ & 8
& 27.20 & 1.80 & \underline{29.00}
& 4.10 & 43.30 & 47.40
& 6.60 & 23.90 & 30.50
& 12.60 & 23.00 & 35.60
& 4.70 & 44.20 & 48.90 \\

RedSage-DPO$^{\star}$ & 8
& \underline{28.30} & 0.60 & 28.90
& \textbf{6.00} & 39.30 & 45.30
& \textbf{9.20} & 22.60 & 31.80
& \underline{14.50} & 20.90 & 35.40
& \textbf{7.20} & 42.70 & 49.90 \\

Qwen3 & 8
& 26.70 & 0.70 & 27.40
& 4.10 & 44.80 & 48.90
& 7.10 & 24.90 & 32.00
& 12.70 & 23.40 & 36.10
& 5.80 & \underline{49.00} & \underline{54.80} \\

Qwen3$^{\dagger}$ & 8
& 25.40 & 0.50 & 25.90
& 4.30 & 45.20 & 49.50
& 6.60 & \underline{26.90} & \textbf{33.50}
& 12.10 & \underline{24.20} & 36.30
& \underline{6.80} & \textbf{53.20} & \textbf{60.00} \\

\midrule

RedSage-Ins$^{\star}$ & 8
& 26.70 & 0.70 & 27.40
& 4.50 & 38.10 & 42.60
& 7.60 & 21.90 & 29.50
& 12.90 & 20.20 & 33.10
& 6.40 & 41.10 & 47.50 \\

\rowcolor{blue!10}
\textbf{RedSage-K (GRPO)$^{\star}$$^{\dagger}$} & 8
& 16.80 & -0.20 & 16.60
& 2.00 & 31.10 & 33.10
& 3.10 & 11.40 & 14.50
& 7.20 & 14.20 & 21.40
& 4.90 & 33.70 & 38.60 \\

\rowcolor{blue!10}
\textbf{RedSage-K (SFT)$^{\star}$} & 8
& 17.80 & 1.20 & 19.00
& 2.40 & 35.30 & 37.70
& 1.50 & 23.60 & 25.10
& 7.30 & 20.00 & 27.30
& 3.80 & 42.70 & 46.50 \\

\rowcolor{blue!10}
\textbf{RedSage-K (SFT+GRPO)$^{\star}$$^{\dagger}$} & 8
& 17.10 & -2.50 & 14.60
& 1.80 & 28.60 & 30.40
& 4.00 & 8.20 & 12.20
& 7.60 & 11.40 & 19.00
& 5.20 & 32.00 & 37.20 \\

\midrule
\multicolumn{17}{>{\columncolor{gray!20}}c}{\textbf{Mid-size Models (20-32B)}} \\
\midrule

Gemma-3-it & 27
& 23.90 & -10.30 & 13.60
& 3.30 & 38.90 & 42.20
& 6.10 & 22.10 & 28.20
& 11.10 & 17.00 & 28.10
& 5.10 & 46.50 & 51.60 \\

GPT-OSS$^{\dagger}$ & 20/3.6
& 22.80 & \textbf{0.80} & \underline{23.60}
& 3.40 & \textbf{45.70} & \textbf{49.10}
& 6.50 & \textbf{28.40} & \textbf{34.90}
& 10.90 & \textbf{24.90} & \textbf{35.80}
& 5.00 & \textbf{53.50} & \underline{58.50} \\

Mistral-Small-3.2 & 24
& \textbf{25.00} & \underline{0.60} & \textbf{25.60}
& 3.90 & \underline{43.30} & 47.20
& \underline{6.60} & \underline{24.00} & \underline{30.60}
& \textbf{11.80} & \underline{22.70} & \underline{34.50}
& 6.70 & \underline{52.50} & \textbf{59.20} \\

Qwen3 & 32
& 22.80 & -0.80 & 22.00
& \textbf{5.50} & 42.30 & \underline{47.80}
& \textbf{7.10} & 23.50 & \underline{30.60}
& \textbf{11.80} & 21.70 & 33.50
& \textbf{7.10} & 52.10 & \textbf{59.20} \\

Qwen3$^{\dagger}$ & 32
& \underline{24.40} & -1.60 & 22.80
& \underline{4.40} & 40.80 & 45.20
& 6.10 & 22.60 & 28.70
& \underline{11.70} & 20.60 & 32.30
& \underline{6.90} & 50.90 & 57.80 \\

\midrule
\multicolumn{17}{>{\columncolor{gray!20}}c}{\textbf{Large-scale Models ($\geq$ 70B)}} \\
\midrule

Llama-Primus-Ins$^{\star}$ & 70
& \underline{23.60} & -4.30 & 19.30
& 4.40 & \textbf{41.80} & \textbf{46.20}
& \underline{7.40} & 23.20 & 30.60
& \textbf{11.80} & 20.20 & 32.00
& 6.20 & 46.20 & 52.40 \\

Llama-3.3-Ins & 70
& \textbf{24.40} & -2.40 & 22.00
& 4.20 & \textbf{41.80} & \underline{46.00}
& 6.20 & \textbf{24.40} & 30.60
& \underline{11.60} & \textbf{21.20} & \textbf{32.80}
& 6.10 & \underline{49.00} & 55.10 \\

GPT-OSS$^{\dagger}$ & 120/5.1
& 22.00 & \underline{0.40} & \textbf{22.40}
& \underline{4.90} & 39.30 & 44.20
& 7.30 & \underline{24.00} & \underline{31.30}
& 11.40 & \textbf{21.20} & \underline{32.60}
& \underline{8.00} & 48.90 & \textbf{56.90} \\

Qwen3-Coder-Next$^{\dagger}$ & 80/3
& 22.70 & -0.40 & \underline{22.30}
& 3.10 & 39.50 & 42.60
& 6.80 & 22.50 & 29.30
& 10.80 & \underline{20.50} & 31.30
& 4.80 & 48.90 & 53.70 \\

Qwen2.5-Ins & 72
& 21.40 & -0.60 & 20.80
& 3.40 & \underline{40.40} & 43.80
& 5.20 & 21.10 & 26.30
& 10.00 & 20.20 & 30.20
& 4.80 & \textbf{50.40} & \underline{55.20} \\

DeepSeek-V3.2$^{\dagger}$ & 685/37
& 22.50 & -1.20 & 21.30
& \textbf{5.00} & 31.50 & 36.50
& 6.60 & 18.70 & 25.30
& 11.40 & 16.30 & 27.70
& 7.20 & 44.30 & 51.50 \\

GLM-5.2$^{\dagger}$ & 753
& 7.90 & \textbf{0.70} & 8.60
& -3.40 & 16.50 & 13.10
& \textbf{27.50} & 11.80 & \textbf{39.30}
& 10.70 & 9.60 & 20.30
& \textbf{10.70} & 22.80 & 33.50 \\

\midrule
\rowcolor{gray!10}
\textbf{Metric Average} & --
& 23.19 & -0.58 & 22.62
& 3.62 & 39.13 & 42.75
& 7.13 & 21.89 & 29.01
& 11.31 & 20.14 & 31.45
& 5.93 & 44.80 & 50.73 \\
\bottomrule

% GLM-5.2$^{\dagger}$ & 753
% & \textbf{41.3} & \textbf{52.0} & 74.8 & \textbf{86.3} & 94.2 & 94.9 & \textbf{75.3} & \textbf{71.9} & 88.4 & 51.3 & \textbf{78.8} & 90.6 & \textbf{71.0} & \textbf{81.7} & 91.3 \\ 

% \midrule
% \rowcolor{gray!10}
% \textbf{Metric Average} & --
%   & - & - & -
%   & - & - & -
%   & - & - & -
%   & - & - & -
%   & - & - & -
% \\

% \bottomrule

\end{tabular}%
}
\end{table}

% \textbf{Metric Average} & --
%   & 22.3 & 28.3 & 73.1
%   & 72.0 & 95.2 & 94.6
%   & 45.1 & 48.7 & 87.8
%   & 61.1 & 68.2 & 90.1
%   & 59.4 & 70.7 & 90.8
% \\

\section{Model Performance Difference across Execution Modes}\label{appen:model_performance}

Table~\ref{tab:deltas} provides a model-wise decomposition of performance changes across execution modes, complementing the raw-score analysis in Table~\ref{tab:combined_perf}. 

A consistent pattern is that improvements in tool selection do not propagate proportionally to exact command correctness. Although Tool $\Delta_1$ is large across all models (23.2 pp on average), Exact $\Delta_1$ remains much smaller (5.9 pp on average). This indicates that even if the tool name can be settled, the remaining errors are still dominated by argument selection, value binding, and command formatting. Thus, restricted mode mainly reduces tool-choice ambiguity, but does not substantially solve command construction.

The restricted-to-hinted transition shows the opposite behavior. Tool $\Delta_2$ is small, unstable, and often negative: 11 out of 24 models show degraded Tool Accuracy after adding full tool manuscripts, with \textit{Gemma-3-it-27B} dropping by 10.3 pp. In contrast, argument-related metrics improve substantially, with Optional F1 increasing by about 39.1 pp and Positional F1 by about 21.9 pp on average. This asymmetry suggests that tool manuscripts are not primarily useful for identifying tools; instead, they help models resolve optional-argument ambiguity, flag selection, value specification, and positional ordering.

Exact Correct further confirms this mechanism. Exact $\Delta_2$ averages 44.8 pp, nearly eight times larger than Exact $\Delta_1$, showing that argument correction is far more effective than tool disambiguation for producing fully correct commands. Model-wise, the largest Exact gains are more closely associated with large Optional $\Delta_2$ than with large Tool $\Delta_1$, suggesting that optional-argument modeling is the key factor determining whether partial command understanding becomes exact command reconstruction.

The \textit{RedSage-K} variants show a distinct delta pattern: despite strong raw performance in the \textit{U} and \textit{R} settings, they benefit less from mode switching than many baselines. In particular, \textit{RedSage-K (GRPO)} and \textit{RedSage-K (SFT+GRPO)} exhibit smaller Optional, Positional, Total, and Exact $\Delta$s than strong general-purpose models. This suggests that KaliBench post-training improves intrinsic command understanding, but the \textit{RedSage-K} variants may be less effective at exploiting full tool documentation, consistent with their weaker \textit{Hinted}-mode gains in the main results.

% Our Kali models show a distinct delta pattern: despite strong raw performance in U and R settings, they gain less from mode switching than many baselines. In particular, Kali-GRPO and SFT+GRPO exhibit smaller Optional, Positional, Total, and Exact $\Delta$s than strong general-purpose models. This suggests that KaliBench post-training improves intrinsic command understanding, but Kali models may be less effective at exploiting long-form tool manuscripts, consistent with their weaker Hinted-mode gains in the main results.

Overall, the delta analysis shows that NL-to-CLI errors are not reduced uniformly across components. The main model-wise differentiator is not the magnitude of tool-selection gains, but how efficiently models use documentation to repair argument errors and produce exact commands.

\begin{figure}[t]
    \centering

    % ----------------- Word Cloud -----------------
    \begin{subfigure}{0.9\columnwidth}
        \centering
        \includegraphics[width=\linewidth]{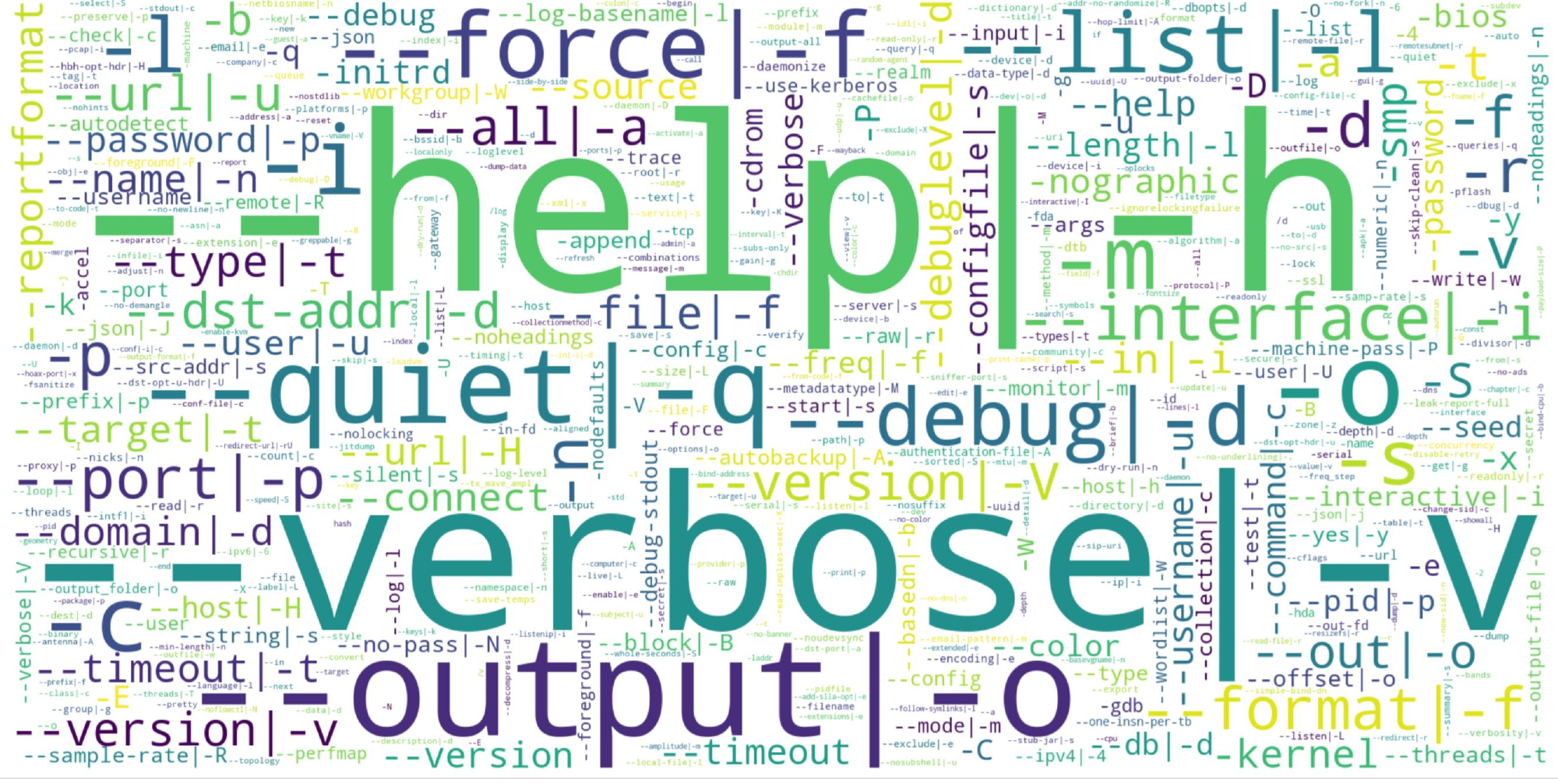}
        \caption{\textbf{Word cloud of options}}
        \label{fig:opt-wordcloud}
    \end{subfigure}

    \vspace{3mm} % spacing between subfigures

    % ----------------- Histogram -----------------
    \begin{subfigure}{0.9\columnwidth}
        \centering
        \includegraphics[width=\linewidth]{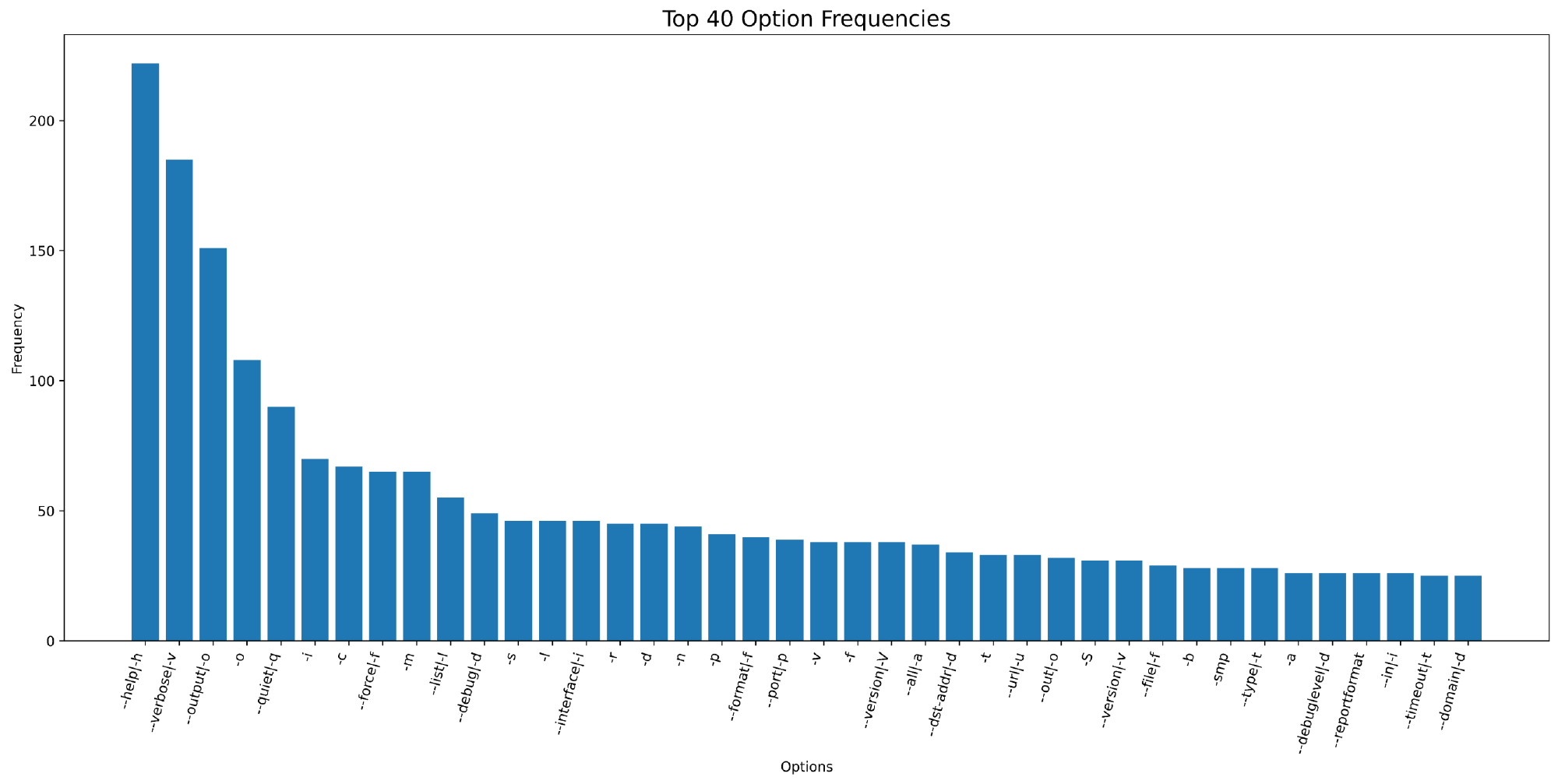}
        \caption{\textbf{Top-40 option frequency distribution.}}
        \label{fig:opt-histogram}
    \end{subfigure}

    \caption{\textbf{Coverage of CLI options in KaliBench (Evaluation Subset).}
   While common flags (e.g., \texttt{--help}, \texttt{--verbose}) appear frequently, many functional tool-specific options also occur across the dataset.}
    \label{fig:opt-coverage}
    \vspace{-3mm}
\end{figure}

\section{RLVR Failure Analysis}
\label{app:rlvr_failure_analysis}
\begin{table}[t]
  \centering
  \caption{
    \textbf{Effect of RLVR on \textit{RedSage-K} across evaluation modes.}
    Scores are Total Score percentages.
    $\Delta$ denotes the change from the original
    \textit{RedSage-Ins} model in percentage points.
}
\label{tab:rlvr_mode_changes}
\small
\setlength{\tabcolsep}{5pt}
\begin{tabular}{lrrrrr}
    \toprule
    \textbf{Mode}
    & \textbf{RedSage-Ins}
    & \makecell{\textbf{RedSage-K} \textbf{(GRPO)}}
    & $\boldsymbol{\Delta}_{\mathrm{\textbf{GRPO}}}$
    & \makecell{\textbf{RedSage-K} \textbf{(SFT+GRPO)}}
    & $\boldsymbol{\Delta}_{\mathrm{\textbf{SFT+GRPO}}}$ \\

      \midrule
      Unrestricted
      & 56.36 & 67.40 & +11.04 & 70.31 & +13.95 \\
      Restricted
      & 69.31 & 74.65 & +5.34 & 77.93 & +8.62 \\
      Hinted
      & 89.53 & 88.77 & -0.76 & 89.34 & -0.19 \\
      \bottomrule
  \end{tabular}
\end{table}

Our paired analysis of all 5,000 evaluation examples across the three settings identifies three recurring RLVR failure modes: invalid optional arguments or option--value pairs, incorrect positional-argument construction, and regressions on examples already solved by the baseline. The model often selects the correct tool but generates incorrect argument forms, such as \texttt{msfvenom --list-formats} instead of \texttt{msfvenom --list formats}.

Although the GRPO and SFT+GRPO variants of \textit{RedSage-K} improve Total Score in U and R, they reduce it in H by 0.76 and 0.19 percentage points, respectively, relative to the original \textit{RedSage-Ins} model. In H, the GRPO variant corrects 537 baseline exact-match failures but regresses on 562 baseline successes. The degradation is concentrated in commands requiring optional arguments or a mixture of optional and positional arguments. 

SFT initialization mitigates these regressions. Compared with GRPO alone, SFT+GRPO improves Total Score by 2.91 points in U and 3.28 points in R, while reducing H-mode exact-match regressions from 562 to 488.

Across security dimensions, both RLVR variants show broad improvements in U and R. We find no robust evidence of a dimension in which RLVR systematically fails to improve; the small variations in H do not form a consistent dimension-specific pattern.

% Our paired analysis of all 5,000 evaluation examples across
% the three settings identifies three recurring RLVR failure
% modes: invalid optional arguments or option--value pairs,
% incorrect positional-argument construction, and regressions
% on examples already solved by the baseline. The model often
% selects the correct tool but generates incorrect argument
% forms, such as \texttt{msfvenom --list-formats} instead of
% \texttt{msfvenom --list formats}.

% Although GRPO and SFT+GRPO improve Total Score in U and R,
% they reduce it in H by 0.76 and 0.19 percentage points,
% respectively, relative to the original RedSage-Ins model.
% In H, GRPO corrects 537 baseline exact-match failures but
% regresses on 562 baseline successes. The degradation is
% concentrated in commands requiring optional arguments or
% a mixture of optional and positional arguments.

% SFT initialization mitigates these regressions.
% Compared with GRPO alone, SFT+GRPO improves Total Score
% by 2.91 points in U and 3.28 points in R, while reducing
% H-mode exact-match regressions from 562 to 488.

% Across security dimensions, both RLVR variants show broad
% improvements in U and R. We find no robust evidence of
% a dimension in which RLVR systematically fails to improve;
% the small variations in H do not form a consistent
% dimension-specific pattern.

\section{Analysis of CLI Option Distribution} \label{distribution}

To assess whether KaliBench captures meaningful CLI behaviors rather than being dominated by generic flags, we analyze option frequencies across evaluation, train, and full sets. In all, there are 3,074 different optional arguments in the evaluation set, 2,739 in the train set, and 4,378 in the full set. Figure~\ref{fig:opt-coverage} shows the top 40 options in the evaluation subset. As expected, broadly supported meta-flags such as \texttt{\dashdash verbose|-v} and \texttt{\dashdash help|-h} appear most frequently. However, the distribution also covers security-relevant CLI patterns, including network and web parameters (\texttt{\dashdash port|-p}, \texttt{\dashdash domain|-d}, \texttt{\dashdash url|-u}), interface/address controls (\texttt{\dashdash interface|-i}, \texttt{\dashdash dst-addr|-d}), and data/output handling flags (\texttt{\dashdash file|-f}, \texttt{\dashdash format|-f}, \texttt{\dashdash output|-o}, \texttt{\dashdash reportformat}).

This distribution confirms that KaliBench does not collapse into trivial command templates. Although generic flags are frequent due to broad tool support, many options correspond to functional behaviors such as scanning configuration, target selection, credential or hash handling, and other core Kali Linux workflows. These security-specific flags demonstrate that KaliBench covers diverse and operationally meaningful command patterns, making it suitable for evaluating fine-grained CLI generation in realistic cybersecurity settings.

% Even when exact scores are lower (e.g., \texttt{spire}, avg\_exact $=0.389$), total scores remain high (avg\_total $=0.824$), indicating that predicting fully canonicalized commands remains challenging due to additive errors accumulated across different stages of CLI formulation.

\section{Complete List of Best-Worst Performed Tools} \label{bestWorst}

Table~\ref{tab:best_worst_by_title} reports representative
high- and low-performing tools after aggregating scores across
24 open-weight model configurations and three evaluation modes.
For each tool, scores are averaged over all associated evaluation
records.

High-performing tools show consistently strong performance
across all sub-metrics. For example, \texttt{d2j-std-apk},
\texttt{rfdump}, and \texttt{truecrypt2john} achieve perfect
average scores, while \texttt{jsql} and \texttt{reconspider}
remain near-perfect with avg\_exact scores of $0.995$ and
$0.993$, respectively. Their high avg\_optional and
avg\_positional scores indicate that the evaluated models
reliably construct arguments for these tools.

By contrast, low-performing tools reveal substantial weaknesses
in argument generation. Although their avg\_total scores range
from $0.347$ to $0.560$, their Exact scores remain much lower.
The clearest example is \texttt{nikto}, which has near-perfect
tool identification (avg\_tool $=0.986$) but zero exact
correctness (avg\_exact $=0.000$), alongside very low
avg\_optional and avg\_positional scores ($0.012$ and $0.042$).
Similar gaps appear for \texttt{steghide}, \texttt{above},
and \texttt{exiflooter}, showing that strong performance on an
individual component is insufficient for complete command
correctness.

\begin{table*}[t!]
  \centering
  \caption{Best- and worst-representative tools aggregated from the scoring files of all models. }
  \label{tab:best_worst_by_title}
  \setlength{\tabcolsep}{3.5pt}
  \begin{tabular}{@{} l
                   S[table-format=1.3]
                   S[table-format=1.3]
                   S[table-format=1.3]
                   S[table-format=1.3]
                   S[table-format=1.3]
                   S[table-format=2.0]
                   S[table-format=1.3] @{}}
    \toprule
    %\rowcolor{blue!10}
    
    \textbf{title}
      & {\textbf{avg\_total}}
      & {\textbf{avg\_tool}}
      & {\textbf{avg\_optional}}
      & {\textbf{avg\_positional}}
      & {\textbf{avg\_exact}}
      & {\textbf{count}}
      & {\textbf{score\_sum}} \\
    \midrule
    \rowcolor{gray!15}
    \multicolumn{8}{c}{\textit{Best Reps}} \\
    \midrule
    d2j-std-apk   & 1.000 & 1.000 & 1.000 & 1.000 & 1.000 & 72 & 5.000 \\
    rfdump & 1.000 & 1.000 & 1.000 & 1.000 & 1.000 & 72 & 5.000 \\
    truecrypt2john    & 1.000 & 1.000 & 1.000 & 1.000 & 1.000 & 72 & 5.000 \\
    jsql    & 0.998 & 1.000 & 0.995 & 1.000 & 0.995 & 216 & 4.989 \\
    reconspider    & 0.998 & 1.000 & 1.000 & 0.993 & 0.993 & 288 & 4.984 \\

    \midrule
    \rowcolor{gray!15}
    \multicolumn{8}{c}{\textit{Worst Reps}} \\
    \midrule
steghide   & 0.560 & 1.000 & 0.347 & 0.333 & 0.319 & 72  & 2.560 \\
above      & 0.490 & 0.747 & 0.357 & 0.368 & 0.319 & 288 & 2.281 \\
exiflooter & 0.491 & 0.719 & 0.379 & 0.375 & 0.191 & 288 & 2.155 \\
hashrat    & 0.464 & 0.524 & 0.273 & 0.593 & 0.184 & 288 & 2.038 \\
nikto      & 0.347 & 0.986 & 0.012 & 0.042 & 0.000 & 72  & 1.386 \\
   
    \bottomrule
  \end{tabular}
  \vspace{1ex}
  \begin{minipage}{\linewidth}
    \small
    \textbf{Notes:} \textit{count} denotes the number of times that a tool appears in the files.
    \textit{score\_sum} is the unweighted sum of all averaged metrics.
  \end{minipage}
\end{table*}

\section{Details about Metrics Calculation} \label{metric_cal}

% NS TODO: Revise the explanation of Optional and Positional F1

\subsection{Command-level Scoring for Evaluating Models}\label{Eva_metric_cal}

For each example, we extract the predicted command from the \texttt{<output>} tag and parse both the ground-truth and predicted commands into: <tool name>, <optional flags>, and <positional arguments> to know which flags take values and which flags are aliases.

\paragraph{Tool score} Tool\_score is a binary indicator: it is 1 if the predicted tool name exactly matches the ground-truth tool name, and 0 otherwise.

\paragraph{Optional-argument F1.}
Ground-truth optional flags are grouped into alias sets (e.g., `-h' | `\verb|--|help'). A predicted flag is counted as correct for a group if it matches any alias in that group. Let $G$ be the number of ground-truth groups, $N_{\text{pred}}$ be the total number of predicted optional flags, and $H$ be the number of ground-truth groups that are ``hit'' by at least one predicted alias. We define $\text{precision}_{\text{opt}} = H / N_{\text{pred}}$ and $\text{recall}_{\text{opt}} = H / G$, and compute
\[
\text{optional\_f1} = \frac{2 \times \text{precision}_{\text{opt}} \times \text{recall}_{\text{opt}}}{\text{precision}_{\text{opt}} + \text{recall}_{\text{opt}}}.
\]
If both ground truth and prediction have no optional flags, we set \textit{optional\_f1} to 1; if ground truth command has no optional flags while the predicted command has, we set it to 0.

\paragraph{Positional-argument F1.}
The ground-truth and predicted positional arguments form 2 sets. We match each predicted argument to an identical ground-truth argument at most once: every successful match counts as a true positive $TP$; any remaining unmatched predicted arguments are false positives $FP$; and any remaining unmatched ground-truth arguments are false negatives $FN$. We then define $\text{precision}_{\text{pos}} = TP / (TP + FP)$ and $\text{recall}_{\text{pos}} = TP / (TP + FN)$, and compute
\[
\text{positional\_f1} = \frac{2 \,\text{precision}_{\text{pos}} \,\text{recall}_{\text{pos}}}{\text{precision}_{\text{pos}} + \text{recall}_{\text{pos}}}.
\]
If both lists are empty, we set \textit{positional\_f1} to 1; if exactly one is empty, we set it to 0.

\paragraph{Total score and Exact match}
Total score is the average of tool score, optional f1 and positional f1. And Exact match evaluates end-to-end command correctness which is 1 for generating a correct canonicalized command, and 0 for any error in parameters.
% \textit{total\_score} is the simple average of \textit{tool\__score}, \textit{optional\_f1}, and positional f1.

% \paragraph{Exact match.}
% \textit{exact\_match} is 1 only if the tool name is correct, the positional-argument lists match exactly (including order), every ground-truth alias group is covered by at least one predicted flag, and no extra flags outside the ground-truth alias sets are predicted; otherwise it is 0.

\subsection{Dimensional Score Calculation} \label{dim_cal}
To analyze model performance across functional security domains, we leverage Kali’s metapackage taxonomy (Section~\ref{kali_tool_dim23}) to assign each tool to one or more of $D=23$ capability dimensions. Each query--command pair is therefore annotated with a multi-hot vector $\mathbf{v}^{(i)} \in \{0,1\}^{D}$ indicating the dimensions to which its ground-truth tool belongs.

For each evaluation instance $i$, we project its scalar prediction score $s^{(i)} \in [0,1]$ onto the corresponding capability dimensions by element-wise multiplication:
\[
\mathbf{s}^{(i)} = s^{(i)} \cdot \mathbf{v}^{(i)}.
\]
This assigns the score $s^{(i)}$ to all dimensions associated with the tool, while all other dimensions receive zero. Summing $\mathbf{s}^{(i)}$ over all instances yields per-dimension performance scores for each model, enabling fine-grained analysis across the 23 cybersecurity capability dimensions.

\paragraph{Example.}
Consider a tool that is associated with three capability dimensions in Kali’s taxonomy, corresponding to dimensions 2, 4, and 6 out of the total $D=23$. Its multi-hot encoding vector is therefore
\[
\mathbf{v}^{(i)} = (0,1,0,1,0,1,0,\ldots,0) \in \{0,1\}^{23},
\]
where the nonzero entries indicate the relevant capability dimensions and all remaining positions are zero.
If the model assigns this instance a scalar prediction score $s^{(i)} = 0.8$, the projected dimensional score vector becomes
\[
\mathbf{s}^{(i)} = 0.8 \cdot \mathbf{v}^{(i)} = (0,0.8,0,0.8,0,0.8,0,\ldots,0).
\]
Thus, the same prediction score is assigned to each of the associated capability dimensions (2, 4, and 6), while having no effect on the others. Aggregating such vectors across all evaluation instances yields cumulative performance scores for each capability dimension.

\section{Detailed Results for Proprietary Models and Agentic Scaffold}
\label{app:propri}

\begin{table}[t]
    \centering
    \caption{
        \textbf{Complete results for proprietary models and agentic scaffolds in the Unrestricted setting.} We used the following abbreviations: Tool Accuracy (TA), Optional-Argument F1 (Opt-F1), Positional-Argument F1 (Pos-F1), Total Score (TS), and Exact Correct (EC). All metric scores are percentages. Bold shows the best.}
    
    \label{tab:proprietary_full}
    \setlength{\tabcolsep}{4pt}
    \renewcommand{\arraystretch}{1.15}
    \resizebox{\textwidth}{!}{%
    \begin{tabular}{@{}lrrrrrrrr@{}}
        \toprule
        \textbf{Model / Scaffold}
        & \makecell{\textbf{Answered}\\\textbf{queries}}
        & \makecell{\textbf{Blocked /}\\\textbf{Missing}}
        & \textbf{TA}
        & \textbf{Opt-F1}
        & \textbf{Pos-F1}
        & \textbf{TS}
        & \makecell{\textbf{EC}\\\textbf{Answered}}
        & \makecell{\textbf{EC}\\\textbf{All 5K}} \\
        \midrule
        GPT-5.6-Sol
        & 4,988 (99.76\%) & 12 (0.24\%)
        & \textbf{92.54} & \textbf{87.42}
        & \textbf{71.51} & \textbf{83.83}
        & \textbf{61.83} & \textbf{61.68} \\

        Claude Opus 5
        & 3,675 (73.50\%) & 1,325 (26.50\%)
        & 90.67 & 85.95 & 68.57 & 81.73 & 59.89 & 44.02 \\

        Codex (GPT-5.5, \texttt{xhigh})
        & 4,633 (92.66\%) & 367 (7.34\%)
        & 91.28 & 82.39 & 63.46 & 79.04 & 55.77 & 51.68 \\
        \bottomrule
    \end{tabular}%
    }
\end{table}

Table~\ref{tab:proprietary_full} reveals a different component-score
pattern from the aggregate open-weight results: both standalone
proprietary models achieve lower Pos-F1 than Opt-F1 despite
high tool-selection accuracy. This pattern also appears in the
Codex configuration, suggesting that argument construction remains
challenging across both standalone and scaffolded settings.
Moreover, Exact Correct (EC) reflects the joint correctness of the entire command
construction process, requiring correct tool selection,
flag--value bindings, and positional arguments. Errors across
these components accumulate, yielding EC scores below the lowest
component score for all three modes. These findings underscore
the value of fine-grained evaluation in explaining the gap between
individual component performance and complete command correctness.

\section{Model-wise Sensitivity to Messy Queries}
\label{app:robust}
We examine whether sensitivity to messy user queries is shared across the evaluated model configurations. Under the Unrestricted setting, we compare each configuration's Total Score on Messy Query inputs with its own performance on the corresponding Clean Query inputs.

\begin{figure*}[t]
    \centering
    \includegraphics[width=\textwidth]{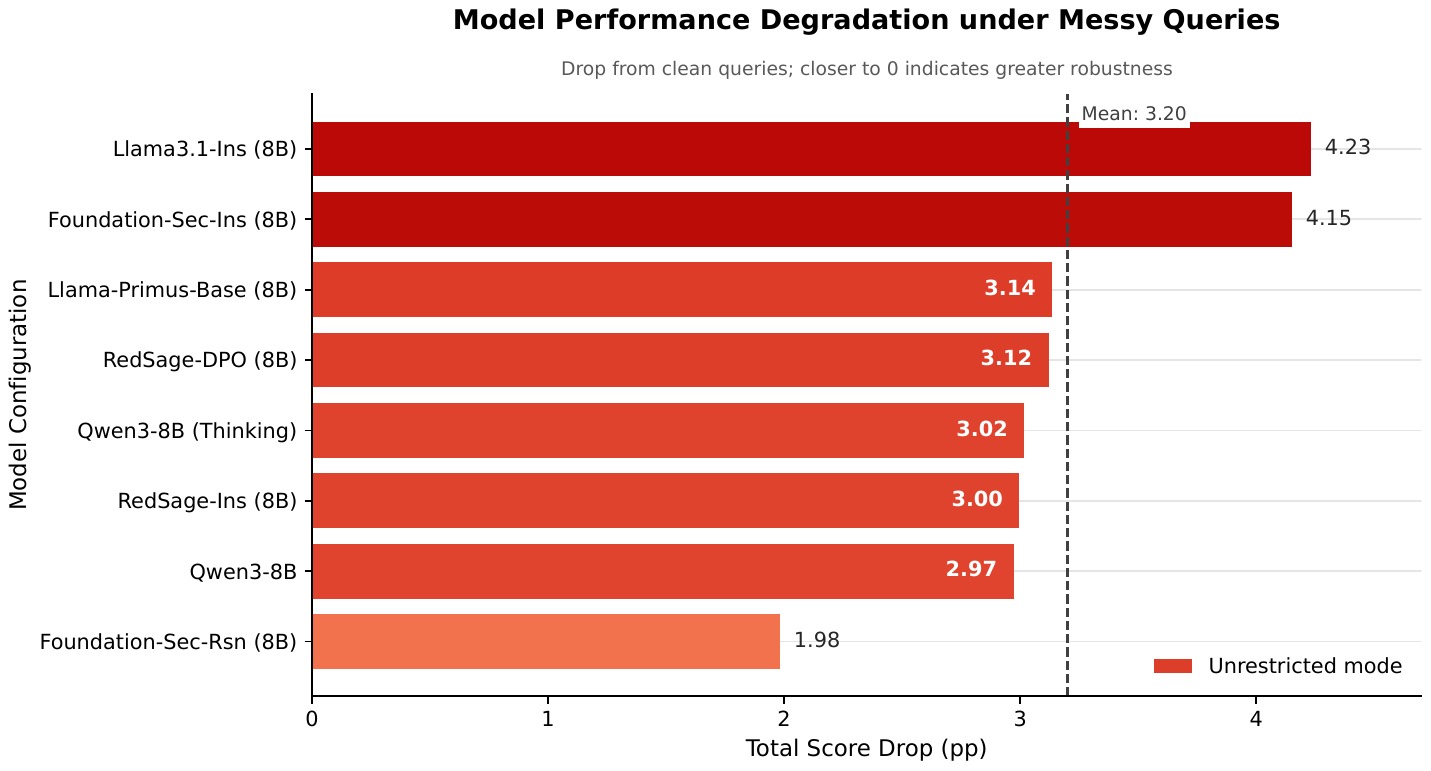}

\caption{
Model-wise Sensitivity to Messy Queries
}
    \label{fig:messyquery}
    \vspace{-1em}
\end{figure*}

Fig~\ref{fig:messyquery} shows that every tested
  configuration experiences a decrease in Total Score.
  Most configurations lose approximately 3 percentage points, while the two most sensitive configurations lose more than 4 percentage points. Even the smallest observed decrease is 1.98 percentage points. This consistent trend across the tested configurations suggests that degradation to messy queries is not confined to a single model family, and how to deal with messy user queries remain a systematic challenge for natural-language-to-CLI translation task. 
%%%%%%%%%%%%%%%%%%%%%%%%%%%%%%%%%%%%%%%%%%%%%%%%%%%%%%%%%%%%

\section{Limitations} \label{limitations}

KaliBench is designed as a deterministic and reproducible benchmark for evaluating NL--to--CLI function calling, and it inherits several limitations intrinsic to the CLI ecosystem and its documentation sources. Ground-truth commands are derived from Kali Linux tool manuals available at dataset construction time. As these manuals evolve across versions through the addition, deprecation, renaming, or reinterpretation of flags, some canonicalized commands may become outdated as upstream documentation changes. In practice, such effects are limited, as tools often maintain backward compatibility, and evaluation in the hinted setting is further insulated from drift by explicitly providing models with the tool documentation to follow.

In addition, real-world tool documentation may occasionally contain inconsistencies or errors. For example, the \texttt{regtree} manual lists the \texttt{-V} option with conflicting semantics (\verb|-|V, \verb|--|no-values and \verb|-|V, \verb|--|version), and several tools in the \textit{Impacket} family document atypical short-option forms (e.g., \texttt{-aesKey}) that differ from common usage conventions (e.g., \verb|--|aesKey). Such cases reflect the diversity and historical evolution of CLI interfaces rather than issues specific to KaliBench, and they make fully deterministic canonicalization inherently approximate when documentation diverges from observed behavior or when multiple variants coexist.

Our LLM-as-Verifier pipeline substantially filters hallucinated flags and arguments by grounding validation in official manuals, but it remains imperfect; human verification of a random subset indicates that residual errors are limited. Similarly, the alias-aware scoring mechanism relies on semi-automatically extracted alias groups and cannot capture all historical, distribution-specific, or community-adopted variants.

Taken together, these considerations position KaliBench as a controlled and well-defined abstraction of real-world CLI tool invocation. It is designed to support systematic, fine-grained evaluation of NL-to-CLI translation, while complementing execution-based or agentic evaluations in operational cybersecurity environments.

\end{document}